\documentclass[11pt]{article}

\usepackage[margin=1in]{geometry}

\usepackage[utf8]{inputenc}
\usepackage[T1]{fontenc}
\usepackage[hypertexnames=false]{hyperref}
\usepackage{url}
\usepackage{booktabs}
\usepackage{etoolbox}
\AtBeginEnvironment{table}{\setlength{\belowcaptionskip}{4pt}}

\usepackage{tabularx}
\usepackage{placeins}
\usepackage{amsfonts}
\usepackage{amsmath}
\usepackage{amssymb}
\usepackage{nicefrac}
\usepackage{microtype}
\usepackage{xcolor}
\usepackage{graphicx}
\usepackage{algorithm}
\usepackage{algorithmic}
\usepackage{multirow}
\usepackage{enumitem}
\usepackage{amsthm}
\usepackage{pgfplots}
\pgfplotsset{
  compat=1.18,
  every axis legend/.append style={
    font=\fontsize{7}{8}\selectfont,
    inner sep=1.5pt,
    nodes={inner xsep=1pt, inner ysep=0.5pt},
    row sep=0pt,
  },
}
\usepackage[numbers,sort&compress]{natbib}

\definecolor{vizblue}{HTML}{0072B2}
\definecolor{vizverm}{HTML}{D55E00}
\definecolor{vizgreen}{HTML}{009E73}

\newtheorem{definition}{Definition}

\newcommand{\framework}{LEGIT}
\newcommand{\tableheading}[1]{{\fontsize{9}{10.5}\selectfont\bfseries #1}}

\newsavebox{\hexpanel}

\title{LEGIT: Credentialing Protocol\\for Trustworthy AI Agent Marketplaces}

\author{%
  Steve Drew\thanks{Corresponding author.}\\
  {\small University of Calgary}\\
  {\small\texttt{steve.drew@ucalgary.ca}}
  \and
  Jiayu Zhou\\
  {\small University of Michigan}\\
  {\small\texttt{jiayuz@umich.edu}}
}

\date{}

\begin{document}

\maketitle

\begin{abstract}
Agentic marketplaces are emerging where AI agents with varying capabilities autonomously complete specialized tasks for buyers.
A major challenge of such marketplaces is that buyers cannot easily determine which agent will perform best on their tasks. Reported benchmark scores may be difficult to verify or compare across tasks, software, and budgets. We introduce \framework{}, a credentialing protocol connecting certification, reputation, and proposed marketplace allocation. Certification binds measured quality and cost per solved task to an agent configuration, task domain, evaluation budget, and evidence through a signed record. Reputation links records of past task outcomes to the same identity, subject to the reliability of the reported feedback. Buyers and agents can verify credential records and inspect optional visual profiles. Evaluations reveal cost differences between agent configurations with similar observed task success, and show that comparisons depend on the evaluation budget. These results support binding performance measurements to the tested configuration and resource limits. A complementary analysis quantifies the deposits and fees required for reputation manipulation under a stated Sybil attack model.
\end{abstract}

\section{Introduction}


AI agents are autonomous systems that combine large language models with software for tool use, memory, and planning. The agent harness is the software around the model that manages prompts, planning, tool calls, memory, and execution. The tool suite specifies which tools are available. Buyers in agent marketplaces choose among competing agents to complete tasks. OpenAI's GPT Store distributes millions of user-built agents~\citep{openai2024gptstore}. Salesforce AgentExchange~\citep{salesforce2025agentexchange} sells agents to enterprises. The AWS Marketplace category for AI agents and tools does the same and launched with more than 900 listings~\citep{aws2025marketplace}. Circle's Agent Stack lets agents browse a marketplace and pay one another in stablecoin~\citep{circle2026agentstack}. Researchers model this agent economy through auction platforms for agent-to-agent value exchange~\citep{yang2025agentexchange} and testbeds for agent labor markets~\citep{liu2026diagon}. They also study economic alignment in multi-agent marketplaces~\citep{karten2026bazaar}. Agent societies organize themselves into productive groups without assigned roles~\citep{pal2026swarmworld}. Traditional software is deterministic and inspectable. Agent capabilities vary between runs~\citep{ouyang2023nondeterminism} and are hard to inspect~\citep{kapoor2024agents}. They depend strongly on harness design choices that buyers cannot see~\citep{yao2024tau}. Buyers who cannot verify quality pay for average quality. This drives good vendors out. Marketplaces therefore need infrastructure that makes quality verifiable.

We examine which evaluation conditions a credential must record for buyers to compare agents. We compare selected models and harnesses on shared tasks at fixed token budgets, then examine retrieval settings and budget sensitivity. A harness comparison measures changes in quality and cost when the surrounding software changes while the model and evaluation conditions are held fixed. The comparisons cover specific configurations, without independently isolating every prompt, memory, or orchestration choice. An agent may also route work among several models or delegate to subagents. Our protocol definition accommodates these dependencies, while our experiments use fixed single-model configurations.

Prior work motivates recording these conditions. On $\tau$-bench, tool-calling harnesses outperform ReAct harnesses by 29\% on the same model~\citep{yao2024tau}. The Efficient Agents framework cuts cost by 28.4\% through harness optimization while retaining 96.7\% of performance~\citep{hu2025efficient}. AgentDiet shows that 39.9--59.7\% of input tokens can be removed without performance loss~\citep{agentdiet2025}. The GAIA leaderboard compares agents with different model and harness configurations~\citep{mialon2023gaia}. These studies already establish the relevance of agent design and resource accounting. LEGIT connects such measurements to a verifiable record that downstream reputation and allocation mechanisms can reference.

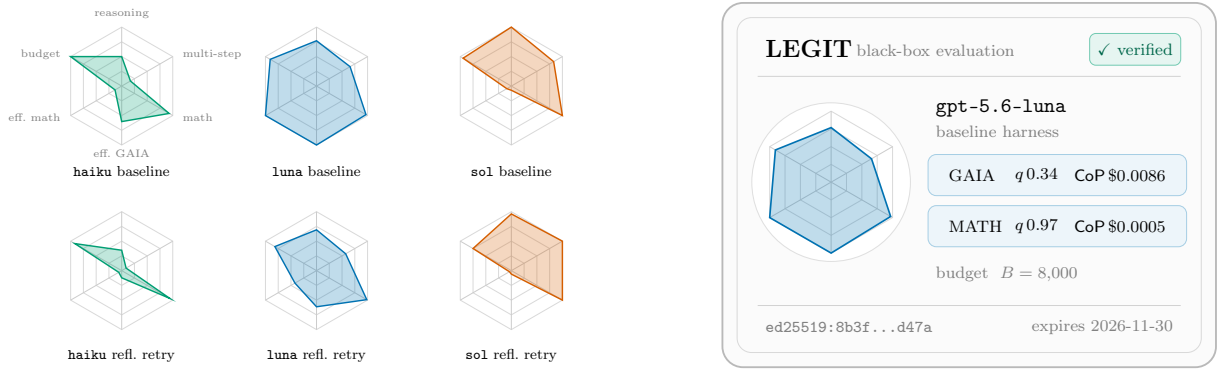
\begin{figure}[!t]
\centering
\sbox{\hexpanel}{%
\newcommand{\hexgrid}{%
  \foreach \r in {0.25,0.5,0.75,1.0} {
    \draw[gray!30, very thin] (90:\r*1.15) -- (30:\r*1.15) -- (-30:\r*1.15)
      -- (-90:\r*1.15) -- (-150:\r*1.15) -- (150:\r*1.15) -- cycle;}
  \foreach \a in {90,30,-30,-90,-150,150} {\draw[gray!40, very thin] (0,0) -- (\a:1.15);}}%
\resizebox{0.45\textwidth}{!}{\begin{tikzpicture}
\begin{scope}[shift={(0,0)}]\hexgrid
\draw[vizgreen, thick, fill=vizgreen, fill opacity=0.25]
  (90:0.575) -- (30:0.197) -- (-30:1.068) -- (-90:0.693) -- (-150:0.150) -- (150:1.150) -- cycle;
\node[font=\scriptsize] at (0,-1.65) {\texttt{haiku} baseline};
\node[font=\tiny, gray] at (90:1.42) {reasoning};
\node[font=\tiny, gray, anchor=west] at (30:1.22) {multi-step};
\node[font=\tiny, gray, anchor=west] at (-30:1.22) {math};
\node[font=\tiny, gray] at (-90:1.32) {eff.\ GAIA};
\node[font=\tiny, gray, anchor=east] at (-150:1.22) {eff.\ math};
\node[font=\tiny, gray, anchor=east] at (150:1.22) {budget};
\end{scope}
\begin{scope}[shift={(3.8,0)}]\hexgrid
\draw[vizblue, thick, fill=vizblue, fill opacity=0.25]
  (90:0.884) -- (30:0.756) -- (-30:1.114) -- (-90:1.150) -- (-150:1.150) -- (150:1.043) -- cycle;
\node[font=\scriptsize] at (0,-1.65) {\texttt{luna} baseline};
\end{scope}
\begin{scope}[shift={(7.6,0)}]\hexgrid
\draw[vizverm, thick, fill=vizverm, fill opacity=0.25]
  (90:1.150) -- (30:0.953) -- (-30:1.150) -- (-90:0.082) -- (-150:0.102) -- (150:1.087) -- cycle;
\node[font=\scriptsize] at (0,-1.65) {\texttt{sol} baseline};
\end{scope}
\begin{scope}[shift={(0,-3.6)}]\hexgrid
\draw[vizgreen, thick, fill=vizgreen, fill opacity=0.25]
  (90:0.398) -- (30:0.099) -- (-30:1.087) -- (-90:0.140) -- (-150:0.066) -- (150:1.055) -- cycle;
\node[font=\scriptsize] at (0,-1.65) {\texttt{haiku} refl.\ retry};
\end{scope}
\begin{scope}[shift={(3.8,-3.6)}]\hexgrid
\draw[vizblue, thick, fill=vizblue, fill opacity=0.25]
  (90:0.796) -- (30:0.657) -- (-30:1.132) -- (-90:0.705) -- (-150:0.486) -- (150:0.938) -- cycle;
\node[font=\scriptsize] at (0,-1.65) {\texttt{luna} refl.\ retry};
\end{scope}
\begin{scope}[shift={(7.6,-3.6)}]\hexgrid
\draw[vizverm, thick, fill=vizverm, fill opacity=0.25]
  (90:1.106) -- (30:1.150) -- (-30:1.150) -- (-90:0.068) -- (-150:0.043) -- (150:0.863) -- cycle;
\node[font=\scriptsize] at (0,-1.65) {\texttt{sol} refl.\ retry};
\end{scope}
\end{tikzpicture}}}
\begin{minipage}{\textwidth}
\begin{minipage}[c]{0.45\textwidth}
\centering
\usebox{\hexpanel}
\end{minipage}\hfill
\begin{minipage}[c]{0.45\textwidth}
\centering
\resizebox{!}{\dimexpr\ht\hexpanel+\dp\hexpanel\relax}{\begin{tikzpicture}
\draw[rounded corners=10pt, fill=gray!3, draw=gray!55, line width=0.9pt]
  (0,0) rectangle (8.4,6.2);
\draw[rounded corners=7pt, draw=gray!30, line width=0.4pt]
  (0.25,0.25) rectangle (8.15,5.95);
\node[anchor=west, font=\large\bfseries] at (0.6,5.42) {LEGIT};
\node[anchor=west, font=\scriptsize, gray] at (2.15,5.40) {black-box evaluation};
\draw[rounded corners=3pt, fill=vizgreen!10, draw=vizgreen!55, line width=0.4pt]
  (6.25,5.14) rectangle (7.8,5.68);
\node[font=\scriptsize, vizgreen!65!black] at (7.025,5.41) {$\checkmark$ verified};
\draw[gray!35, line width=0.4pt] (0.6,4.95) -- (7.8,4.95);
\draw[gray!25, fill=white, line width=0.4pt] (1.85,3.15) circle (1.35);
\begin{scope}[shift={(1.85,3.15)}, scale=1.05]
  \foreach \r in {0.25,0.5,0.75,1.0} {
    \draw[gray!30, very thin] (90:\r*1.15) -- (30:\r*1.15) -- (-30:\r*1.15)
      -- (-90:\r*1.15) -- (-150:\r*1.15) -- (150:\r*1.15) -- cycle;}
  \foreach \a in {90,30,-30,-90,-150,150} {\draw[gray!40, very thin] (0,0) -- (\a:1.15);}
  \draw[vizblue, thick, fill=vizblue, fill opacity=0.25]
    (90:0.884) -- (30:0.756) -- (-30:1.114) -- (-90:1.150) -- (-150:1.150) -- (150:1.043) -- cycle;
\end{scope}
\node[anchor=west, font=\small] at (3.50,4.42) {\texttt{gpt-5.6-luna}};
\node[anchor=west, font=\scriptsize, gray] at (3.50,4.02) {baseline harness};
\draw[rounded corners=4pt, fill=vizblue!7, draw=vizblue!35, line width=0.4pt]
  (3.50,2.92) rectangle (7.90,3.62);
\node[anchor=west, font=\scriptsize] at (3.72,3.27) {GAIA};
\node[anchor=west, font=\scriptsize] at (4.85,3.27) {$q\,0.34$};
\node[anchor=east, font=\scriptsize] at (7.68,3.27) {$\mathsf{CoP}\,\$0.0086$};
\draw[rounded corners=4pt, fill=vizblue!7, draw=vizblue!35, line width=0.4pt]
  (3.50,2.05) rectangle (7.90,2.75);
\node[anchor=west, font=\scriptsize] at (3.72,2.40) {MATH};
\node[anchor=west, font=\scriptsize] at (4.85,2.40) {$q\,0.97$};
\node[anchor=east, font=\scriptsize] at (7.68,2.40) {$\mathsf{CoP}\,\$0.0005$};
\node[anchor=west, font=\scriptsize, gray] at (3.50,1.58) {budget \ $B = 8{,}000$};
\draw[gray!35, line width=0.4pt] (0.6,1.05) -- (7.8,1.05);
\node[anchor=west, font=\scriptsize, gray] at (0.6,0.68)
  {\texttt{ed25519:8b3f\ldots d47a}};
\node[anchor=east, font=\scriptsize, gray] at (7.8,0.68) {expires 2026-11-30};
\end{tikzpicture}}
\end{minipage}
\end{minipage}
\caption{Example views of structured agent credentials. Left, normalized profiles compare quality, efficiency, and budget sensitivity. Right, a card displays measured \texttt{gpt-5.6-luna} baseline scores with illustrative signature and expiry fields. The visual verification mark is not itself a verification result. Section~\ref{sec:credential-profiles} defines the axes and fields.}
\label{fig:hex}
\label{fig:card}
\end{figure}

LEGIT expresses an agent's credential as a signed, structured profile of its measured quality, cost, configuration, and evaluation conditions. Figure~\ref{fig:hex} illustrates optional visual views of that record. The credential remains machine-verifiable without either view. Evaluation dates and expiry delimit its temporal scope, while a history of evaluations can expose changes in performance.

\paragraph{Contributions.}

\begin{enumerate}[leftmargin=*, itemsep=2pt]
\item We formalize the problem of verifiable agent credentials. We present \framework{}, a three-layer protocol with definitions of quality, cost, reputation, and auction scores.
\item We specify a structured credential profile that binds domain quality and cost to an agent configuration, evaluation budget, evidence, and expiry. An issuer signature supports verification. Charts and cards illustrate ways to present the same record.
\item We compare eighteen agent configurations under controlled harness and budget settings. Within-model cost differences and budget-sensitive retrieval rankings motivate reporting configuration, quality, cost, and budget together. Corrected paired analyses support model-quality differences, while harness-quality differences remain inconclusive in the tested grid.
\item We assess economic barriers to reputation manipulation by quantifying locked capital and spent fees under a Sybil attack model. A Sybil identity here is an attacker-controlled identity used alongside others to accumulate misleading reputation. The analysis compares funding requirements under stated deposit, fee, and feedback assumptions.
\end{enumerate}

\FloatBarrier

\section{Related Work}

Existing research provides methods to evaluate agents, improve their workflows, document their capabilities, and allocate work. Buyers need these pieces connected so that measured performance can guide task selection. LEGIT links evaluation evidence to an agent's credentials and reputation, then uses those records in proposed marketplace allocation.

\subsection{Agent Benchmarks}

Agent benchmarks test different kinds of work. AgentBench evaluates language models across eight interactive environments~\citep{liu2023agentbench}. GAIA combines reasoning, web retrieval, tool use, and multimodal information in 466 questions for general assistants. It uses normalized exact-match scoring across three difficulty levels~\citep{mialon2023gaia}. SWE-bench evaluates changes to software repositories using real GitHub issues and executable tests~\citep{jimenez2024swebench}. SWE-bench Verified provides a human-validated subset~\citep{openai2024sweverified}. These benchmarks make task completion observable under specified evaluation rules.

WebArena supplies reproducible websites and checks functional completion across 812 tasks~\citep{zhou2024webarena}. VisualWebArena adds tasks that require visual understanding~\citep{koh2024visualwebarena}. WorkArena focuses on enterprise workflows, while AssistantBench studies realistic information needs on the open web~\citep{drouin2024workarena,yoran2024assistantbench}. BrowserGym standardizes interactions and experiment management across web benchmarks~\citep{lesellier2024browsergym}. OSWorld extends execution-based evaluation to desktop applications and operating system tasks~\citep{xie2024osworld}.

Customer service also requires agents to follow policies and coordinate with users. $\tau$-bench measures tool use and repeated-trial reliability in this setting. Its reliability metric counts a task as solved only when every repeated trial succeeds~\citep{yao2024tau}. $\tau^2$-bench allows both the user and agent to change a shared environment, so that success also depends on communication and coordination~\citep{barres2025tau2bench}.

A benchmark score describes performance within its own tasks and evaluation rules. Buyers still need a record of the agent configuration and resources behind that score. LEGIT addresses this gap by binding quality and cost measurements to a configuration, task domain, and token budget. Its credentials make the scope of a comparison explicit.

\subsection{Evaluation Credibility and Reliability}

HELM evaluates language models under standardized conditions using accuracy, efficiency, robustness, and other metrics~\citep{liang2022helm}. AI Agents That Matter argues for evaluating agent accuracy and cost together and identifies problems with reproducibility and benchmark design~\citep{kapoor2024agents}. The Holistic Agent Leaderboard (HAL) provides shared evaluation infrastructure, compares models and harnesses across benchmarks, and publishes execution traces~\citep{kapoor2025hal}. Joint quality and cost evaluation is therefore an established foundation for agent assessment.

Reported rankings can also depend on how evidence is collected and disclosed. The Leaderboard Illusion documents selective reporting and uneven private testing in Chatbot Arena~\citep{singh2025leaderboard}. Oren et al. detect test-set contamination through differences between original and shuffled dataset orderings under exchangeability assumptions~\citep{oren2023contamination}. Jacovi et al. propose practices for distributing benchmark data that reduce its exposure to training pipelines~\citep{jacovi2023stop}. LiveBench uses regularly updated questions and objective scoring to limit contamination and judging bias~\citep{white2025livebench}.

Reliability also extends beyond average task success. Rabanser et al. evaluate consistency, robustness, predictability, and safety~\citep{rabanser2026reliability}. Kwa et al. relate agent success to the time humans need for software tasks, giving an interpretable measure of task difficulty~\citep{kwa2025longtasks}.

These methods improve the evidence available to buyers. A marketplace also needs to preserve the link between that evidence and the agent offered for work. LEGIT addresses this reporting gap through credentials that record the tested configuration and evaluation conditions. The same record supports visual comparison, reputation tracking, and proposed allocation. Benchmark integrity remains a property of the evaluation evidence on which the credential relies.

\subsection{Agent Harnesses and Workflow Design}

An agent's behavior depends on the software around its model. ReAct interleaves reasoning with actions that obtain information from an environment~\citep{yao2022react}. Reflexion stores feedback as verbal reflections for later attempts~\citep{shinn2023reflexion}. Language Agent Tree Search combines search, reflection, and feedback when selecting actions~\citep{zhou2023lats}. These methods change how an agent spends computation and uses its observations.

Toolformer trains models to decide when and how to call tools~\citep{schick2023toolformer}. ToolLLM combines API instruction data, training, and tool-use evaluation~\citep{qin2023toolllm}. Gorilla combines API-call training with retrieval of tool documentation~\citep{patil2023gorilla}. SWE-agent shows that the interface for navigating and editing repositories affects software task performance~\citep{yang2024sweagent}. Tool access and interface design therefore belong in the description of the evaluated agent.

Collaboration and workflow optimization introduce further choices. AutoGen provides programmable conversations among agents, tools, and people~\citep{wu2023autogen}. MetaGPT organizes collaboration through roles and operating procedures~\citep{hong2023metagpt}. ChatDev coordinates software development through dialogue among specialized agents~\citep{qian2023chatdev}. DSPy optimizes modular language-model programs against a selected metric~\citep{khattab2023dspy}. AFlow searches over workflows using execution feedback, while Automated Design of Agentic Systems generates and evaluates agent programs~\citep{zhang2024aflow,hu2024adas}.

These systems help developers construct and improve agents. Their construction methods do not by themselves tell a buyer which configuration produced a reported score. LEGIT addresses this identification gap by treating the model dependencies, harness, tools, and parameters as the unit of certification. The credential associates measured performance with that unit.

\subsection{Cost and Resource Allocation}

Agent optimization already accounts for the trade-off between quality and expense. FrugalGPT uses model cascades to reduce query cost while maintaining response quality~\citep{chen2023frugalgpt}. RouteLLM learns to select models from preference data~\citep{ong2024routellm}. LLMCompiler plans dependencies among tool calls and executes independent calls concurrently to reduce latency and cost~\citep{kim2023llmcompiler}.

AgentDiet reduces token use by removing redundant and outdated information from execution histories while maintaining performance in its evaluations~\citep{agentdiet2025}. CodeAgents represents multi-agent interactions as structured pseudocode~\citep{codeagents2025}. Its HotpotQA ablation reports a 26.8\% relative accuracy gain with 60\% fewer tokens than its natural-language baseline without replanning. Efficient Agents examines model choice, agent design, and test-time computation using cost per successful solution~\citep{hu2025efficient}. The syftr framework searches for workflows that offer favorable combinations of accuracy and cost~\citep{syftr2025}. These studies provide methods for finding economical configurations and comparing their outcomes.

An optimized configuration still needs measurements that buyers can interpret under their own task constraints. LEGIT addresses this comparison gap by recording quality and cost per solved task alongside the evaluation budget. Its experiments examine how harness rankings change across budgets and how costs differ when task success is comparable. These measurements supply the credential fields used in the proposed marketplace score.

\subsection{Performance Reports and Verifiable Credentials}

Model Cards describe intended uses, evaluation procedures, and performance under different conditions~\citep{mitchell2019modelcards}. Datasheets document dataset composition, collection, and recommended uses~\citep{gebru2021datasheets}. FactSheets extend structured disclosure to AI services, including performance and provenance~\citep{arnold2019factsheets}. These reporting approaches help users interpret the evidence behind a system's reported capabilities.

Identity standards provide a way to associate claims with their subject and issuer. The W3C Decentralized Identifiers standard specifies identifiers and methods for proving control over them~\citep{w3c2022did}. The Verifiable Credentials data model defines how issuers express claims that holders can present to verifiers~\citep{w3c2025vc}. Rodriguez Garzon et al. apply these mechanisms to AI agents and demonstrate exchanges of credentials bound to agent identities~\citep{rodriguezgarzon2025ai}.

General reporting formats and identity standards leave the choice of task-performance fields to their application. LEGIT specifies domain-specific quality, cost, and budget fields linked to an agent configuration and evaluation evidence. An issuer signs the record, which automated verifiers can inspect. Visual profiles and cards are example presentations of these fields. The contribution is the scoped measurement record and its connection to reputation and proposed allocation.

\subsection{Reputation and Sybil Identities}

The Beta Reputation System accumulates positive and negative feedback to estimate reputation~\citep{ismail2002beta}. J{\o}sang and Quattrociocchi develop features for sparse ratings, initial beliefs, and changing behavior~\citep{josang2009bayesian}. EigenTrust combines peers' local assessments into reputation values for selecting file providers~\citep{kamvar2003eigentrust}. These approaches connect past interactions to later decisions about service providers.

Multiple identities controlled by one attacker complicate this connection. Douceur shows how one participant can undermine a system by presenting multiple identities~\citep{douceur2002sybil}. Friedman and Resnick analyze the consequences of cheap replacement identities and the trade-offs introduced by entry fees~\citep{friedman2001pseudonyms}. Identity creation and reputation accumulation must therefore be considered together.

These general models leave the funding requirements of LEGIT's reputation rule unspecified. LEGIT combines a discounted Beta update with a deposit for each new identity. Its analysis separates locked capital from spent interaction fees and computes the resources needed to reach a target reputation. The resulting estimates let operators compare deposit, fee, and reputation settings under the modeled feedback conditions.

\subsection{Agent Marketplaces and Procurement}

Scoring auctions give buyers a way to consider quality alongside price. Che studies procurement in which firms bid on both dimensions~\citep{che1993design}. Asker and Cantillon analyze scoring auctions with multiple private attributes~\citep{asker2008scoring}. Papakonstantinou and Bogetoft use payments after observing delivered quality to create incentives for truthful reporting in their procurement model~\citep{papakonstantinou2013incentives}. These mechanisms require a defined meaning for the quality attributes used in allocation and payment.

Agent Exchange proposes auction infrastructure that connects capability representation, performance tracking, and agent coordination~\citep{yang2025agentexchange}. Diagon provides configurable experiments on allocation, contracting, and enforcement in agent labor markets~\citep{liu2026diagon}. Agent Bazaar studies market stability and deception by sellers controlling multiple identities~\citep{karten2026bazaar}. AgentSLA supplies a quality model and a language for specifying service agreements for agents~\citep{jouneaux2025agentsla}.

Mittal's Trust Layer is particularly close to LEGIT. It combines capability descriptors, screening, and reputation to address unreliable capability advertisements~\citep{mittal2026capability}. Its analysis uses equilibrium arguments and illustrative simulations with stipulated provider reliability. LEGIT adds controlled measurements of model, harness, and budget effects to motivate the contents of a quality and cost credential.

Marketplace mechanisms and service agreements need comparable evidence for the attributes on which they act. LEGIT addresses this interface through a credential with explicit configuration, domain, and budget scope. Its proposed allocation rule reads certified quality and cost together with reputation and bid price. This connects measured agent performance to eligibility and allocation decisions.

\section{Problem Formulation}


\subsection{System Model}

A marketplace $\mathcal{M}$ has agent vendors $\mathcal{V} = \{v_1, \ldots, v_m\}$ who register agents. Task requesters $\mathcal{R} = \{r_1, \ldots, r_n\}$ post task specifications. Certification authorities $\mathcal{C} = \{c_1, \ldots, c_p\}$ issue and verify credentials. Let $\mathcal{D}$ denote the task domains, such as code generation, customer service, or research synthesis. Let $k = |\mathcal{D}|$ denote their number.

\begin{definition}[Agent]
An agent $a = (M, S, T, P)$ contains a collection of model dependencies $M$, an agent harness $S$, a tool suite $T$, and a parameter configuration $P$. The collection $M$ may contain one model or several models used through routing or delegation. The harness contains the system prompt, planning and orchestration logic, tool-call control, memory architecture, and model-selection policy. The tool suite identifies the available tools and delegated services. The configuration sets temperature, sampling strategy, and context management. A single-model agent is the special case with one model dependency.
\end{definition}

\begin{definition}[Task Specification]
A task specification $\tau = (\delta, B, Q_{\min}, D)$ contains a task description $\delta$ and a maximum token budget $B \in \mathbb{N}$. It also sets a minimum quality threshold $Q_{\min} \in [0,1]$ for the task's domain and a domain label $D \in \mathcal{D}$.
\end{definition}

\begin{definition}[Agent Credential]
An agent credential is $\Gamma_a=(\xi_a,\mathbf{q}_a,\mathbf{e}_a,\pi_a,t_a,\sigma_a)$. The scope record $\xi_a$ identifies the agent and its configuration $a=(M,S,T,P)$, including the disclosed model versions and routing or delegation policy. It distinguishes evaluator-observed configuration details from vendor declarations and unknown dependencies. It records the certified domains $\mathcal{D}_{\mathrm{cert}}$, each domain's task set or sampling record, scoring rule, and token budget $B_D$. It also records the currency, usage prices, evaluation date, and issuer. The vectors $\mathbf{q}_a$ and $\mathbf{e}_a$ have one entry per certified domain. Their entries give mean quality and cost per solved task. An omitted domain is unevaluated. A zero quality score means that the domain was evaluated but no task earned credit. Its cost per solved task is infinite. The evidence record $\pi_a$ links the credential to recorded outputs, scores, and usage. The timestamp $t_a$ sets expiry. The issuer signature $\sigma_a$ covers the scope, scores, evidence record, and expiry. Compact architecture labels use $q_a$ and $e_a$ for the vectors and $\pi$ for $\pi_a$.
\end{definition}

Layer~2 maintains the reputation score $\rho_a \in [0,1]$, as defined in Section~\ref{sec:layer2}. The score is bound to the same agent identity. Layer~3 uses it alongside $\Gamma_a$. Reputation changes with every task outcome, so it is excluded from the static Layer~1 credential signed at certification time.

The credential is a signed data record. A content hash identifies the record or linked evidence, while the issuer signature authenticates the signed claim under the verifier's trusted issuer policy. Neither a hash nor a signature proves that an undisclosed model dependency matches a vendor declaration. For opaque services, the scope identifies the tested service, its declared configuration, and the available observations. The resulting assurance is limited to that evidence.

\subsection{Evaluation Questions}

We ask whether selected model and harness configurations differ in task success under the same token budget. Comparisons share tasks to control task difficulty. For the omnibus model and harness checks, the null assumes that the tested labels are exchangeable within each task. For an exact paired comparison of binary outcomes, the null assigns equal probability to either configuration solving a task that only one solves. The tests address quality differences within the evaluated configuration set.

We also ask how cost per solved task and rankings vary with configuration, domain, and budget. Paired-bootstrap intervals summarize uncertainty in reported quality and cost ratios. A ratio of 1 denotes equal values of the compared metric. Saved-run threshold curves describe budget sensitivity, with a separate direct-run check. These analyses motivate the fields included in a credential. They do not statistically test the necessity of the protocol or its effect on marketplace outcomes.

\subsection{Evaluation Scope}

The experiments cover three models from two vendors, selected harnesses and retrieval settings, and task subsets from GAIA and MATH-500. Each evaluated configuration uses one model. Harness variants combine design choices, and the extended grid includes runs from a later round with a different invocation-cache setting. These comparisons do not isolate every component or establish general results for dynamic model mixtures, delegated agents, or other domains. Reputation tracks later task outcomes subject to their reporting and verification assumptions.

The Sybil analysis calculates the deposits and fees needed to accumulate reputation under the attack assumptions in Section~\ref{sec:threats}. End-to-end allocation benefits, strategic bidding behavior, and long-term credential drift are not evaluated. A future allocation study could compare credential-based selection with model-only, quality-only, and price-only baselines on held-out tasks, measuring delivered quality, cost, and selection regret.

\begin{figure}[!t]
\centering
\includegraphics[width=0.98\textwidth]{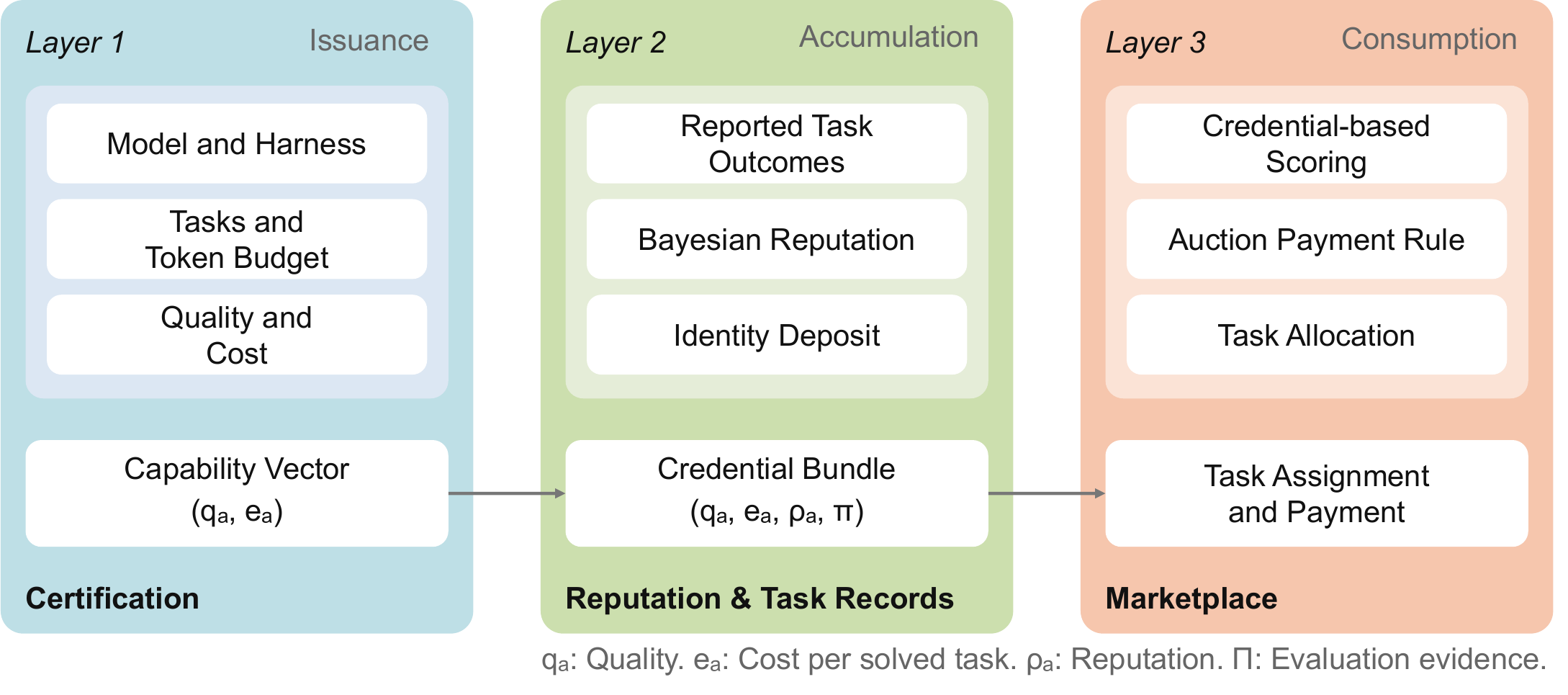}
\caption{The LEGIT architecture connects certification, reputation records, and marketplace allocation. Credentials carry measured quality and cost into reputation tracking, task allocation, and payment.}
\label{fig:overview}
\end{figure}

\section{The \framework{} Protocol}


Figure~\ref{fig:overview} links capability measurement, reputation records, and task allocation.

\subsection{The Certification Layer}


\paragraph{Capability evaluation.} Layer~1 issues capability vectors by evaluating agents on benchmark tasks under fixed token budgets.

\paragraph{Certification scope.} Each certification authority defines a catalog of domains $\mathcal{D}$ with associated task distributions and quality metrics. For example, a code generation domain might count a task as solved when the first attempt clears its unit tests. A customer service domain might use task completion rate as in $\tau$-bench~\citep{yao2024tau}. A research synthesis domain might use factual accuracy scored by human evaluators. The quality function $Q(a, \tau, B) \in [0,1]$ is thus domain specific. The certification authority fixes its definition and publishes it alongside the domain specification. Both agents and requesters then know what is being measured. An agent selects which domains to certify in based on its intended market. It receives scores only for those domains. The certification ecosystem can therefore grow as new task categories emerge without changes to the protocol itself.

\paragraph{Evaluation budget.} Each task is evaluated against a token allowance $B_D$ for its domain. Let $\mathcal{D}_{\mathrm{cert}}$ contain the domains selected for certification. For each domain $D$, let $\mathcal{T}_D$ be its evaluation task set. Let $C(a,\tau)$ be the monetary cost of executing agent $a$ on task $\tau$. The quality score $Q(a,\tau,B_D)$ lies between 0 and 1. Over-budget tasks receive zero credit.

Following~\citet{hu2025efficient}, cost per solved task divides total cost by total credited quality. For binary scores, total credited quality is the number of solved tasks. Costs include failed attempts.

\begin{equation}
\mathsf{CoP}(a,\mathcal{T}_D,B_D)
= \frac{\sum_{\tau\in\mathcal{T}_D} C(a,\tau)}
       {\sum_{\tau\in\mathcal{T}_D} Q(a,\tau,B_D)}.
\end{equation}
When no task earns credit, cost per solved task is recorded as infinite. A lower value means less money spent per unit of credited quality. The domain credential contains mean quality and cost per solved task.

\begin{equation}
q_{a,D}=\frac{1}{|\mathcal{T}_D|}\sum_{\tau\in\mathcal{T}_D}Q(a,\tau,B_D),
\qquad e_{a,D}=\mathsf{CoP}(a,\mathcal{T}_D,B_D),
\qquad D\in\mathcal{D}_{\mathrm{cert}}.
\end{equation}

The certified credential for domain $D$ is the pair $(q_{a,D},\, e_{a,D})$. It captures what the agent achieves and what it costs. Neither metric alone suffices. Agents with identical $\mathsf{CoP}$ may differ substantially in absolute quality. Agents with identical quality may differ by an order of magnitude in cost. Each task requester can choose a trade-off between quality and cost through the Layer~3 scoring rule. Task requesters can filter agents by quality thresholds and $\mathsf{CoP}$ ceilings. Certification authorities can deny credentials to agents whose efficiency falls below minimum standards. This two-dimensional credential serves as the agent's profile in downstream layers. Layer~2 associates reported deployment outcomes with the same identity. Layer~3 uses it to filter eligible bidders and weight auction scores.

\paragraph{Credential lifecycle.} A credential describes an evaluation at a recorded time. Its expiry limits acceptance of that record but does not guarantee unchanged behavior until expiry. The issuer or verifier can impose a maximum acceptable age. A change to the declared model dependencies, routing policy, harness, tools, or parameters changes the certified configuration and calls for a new evaluation before the revised agent is represented as covered by that credential. An unchanged routing policy may select different declared models across tasks, provided this behavior lies within the evaluated scope. Undisclosed provider changes remain an assurance limitation for opaque services.

We propose retaining immutable historical records and linking replacement credentials to the records they supersede. A revoked or suspended record should be rejected for current eligibility even if its signature remains valid. Comparisons over time should retain task-set or sampling provenance, scoring rules, budgets, and price snapshots. Changes in quality, cost, and uncertainty can then be reported with any change in evaluation conditions. The implementation supports signed manifests, issuance and expiry timestamps, credential status, and policy-based verification. Automatic detection of configuration drift, links between superseding records, and longitudinal performance analysis are proposed lifecycle extensions, not evaluated features.

\subsection{The Reputation and Portfolio Layer}
\label{sec:layer2}

\paragraph{Task records.} Layer~2 associates records of completed tasks and a reputation score with the same agent identity.

\paragraph{Outcome provenance.} The reputation model updates its score from records of past task outcomes. The quality signal is an input supplied by the marketplace's designated reporting or evaluation process, which may use requester feedback, an automated task check, or an evaluator's assessment. The update equation does not verify the signal. A deployment should record the task, agent configuration, reporter, scoring rule, time, and supporting evidence, and distinguish independently checked outcomes from unverified feedback. The analysis below assumes the stated feedback process without implementing a general dispute-resolution or feedback-authentication mechanism.

\paragraph{Agent identity.} Each agent has an identifier that links its capability credential, task records, and reputation score.

For an agent that changes configuration, task records should retain the configuration used at execution time. A deployment should expose that history and state whether reputation is carried across versions, so buyers can distinguish earlier performance from outcomes under the current configuration. This version-aware policy is a proposed extension of the analyzed identity-level update.

\paragraph{Bayesian reputation model.} For each quality dimension $d$, we maintain a Beta distribution posterior over the agent's true quality parameter $\theta_d$, so that

\begin{equation}
p(\theta_d | \text{data}) = \mathsf{Beta}(\alpha_d, \beta_d)
\end{equation}

The positive Beta parameters $\alpha_d$ and $\beta_d$ represent accumulated evidence of success and failure for dimension $d$. In the analyzed model, a new identity starts with $\alpha_d^{(0)}=\beta_d^{(0)}=1$, giving a uniform prior. The index $t$ counts completed task outcomes, starting at 0 before any feedback. Following J{\o}sang and Quattrociocchi~\citep{josang2009bayesian}, we incorporate a longevity factor $\eta \in (0, 1]$ that geometrically discounts older observations, giving

\begin{equation}
\alpha_d^{(t)} = \eta \cdot \alpha_d^{(t-1)} + s_d, \quad \beta_d^{(t)} = \eta \cdot \beta_d^{(t-1)} + (1 - s_d)
\end{equation}

where $s_d \in [0,1]$ is the quality signal for dimension $d$ from the most recent interaction.

The reputation score for agent $a$ on dimension $d$ is the posterior mean $\rho_{a,d}^{(t)}=\alpha_d^{(t)}/(\alpha_d^{(t)}+\beta_d^{(t)})$. The parameters belong to agent $a$, whose index is omitted from the update equations. In a task's marketplace score, $\rho_a$ denotes this reputation for the task's rated dimension. The Sybil analysis considers one dimension and omits the agent and dimension indices.

\paragraph{Sybil funding requirements.} The modeled Sybil attack uses multiple attacker-controlled identities to accumulate misleading reputation. These identities may have valid identifiers and keys. The attack does not require forging another party's signature or identity. Let $N_{\mathrm{Sybil}}$ count the attacker-controlled identities. Each requires a deposit $s_{\min}$, so the attacker must lock at least $N_{\mathrm{Sybil}}s_{\min}$ in capital. Section~\ref{sec:sybil} separates locked capital from spent interaction fees and states the limits of this economic barrier.

\subsection{The Marketplace Layer}


\paragraph{Task allocation.} Layer~3 proposes a use of the measured credentials in task allocation. Its scoring rule combines quality, cost, reputation, and price.

\paragraph{Bid structure.} The marketplace filters eligible agents for task specification $\tau = (\delta, B, Q_{\min}, D)$. It uses Layer~1 credentials $(\mathbf{q}_a, \mathbf{e}_a)$ and Layer~2 reputation $\rho_a$. An eligible agent $a$ bids a quality commitment, a price, and a credential record. The quality commitment $q_a^{\text{commit}} \in [Q_{\min},\, q_{a,D}]$ is the quality level the agent pledges to deliver. It is bounded below by the task's minimum threshold and above by the agent's certified quality for domain $D$. The price $p_a$ is the monetary compensation the agent requests. The credential record $\sigma_a^{\text{cert}}$ contains the signed Layer~1 credential and its linked Layer~2 reputation record. The operator checks the issuer signature, configuration and evaluation scope, expiry, applicable age policy, and current status. A revoked or suspended credential is ineligible. The marketplace reads certified efficiency $e_{a,D}$ directly from the credential. Agents do not bid on it. The bid tuple is

\begin{equation}
b_a = (q_a^{\text{commit}}, p_a, \sigma_a^{\text{cert}})
\end{equation}

\paragraph{Scoring rule.} Following Che\textquotesingle s multi-dimensional scoring framework~\citep{che1993design}, we define the proposed score

\begin{equation}
\mathsf{Score}(b_a) = w_q \cdot q_a^{\text{commit}} + w_e \cdot \left(1 - \frac{e_{a,D}}{e_{\max}}\right) + w_\rho \cdot \rho_a - w_p \cdot \frac{p_a}{p_{\max}}
\end{equation}

The limits $e_{\max}>0$ and $p_{\max}>0$ are the maximum eligible cost per solved task and bid price. The credential must cover the requested domain and budget, so $B_D=B$. An eligible bid has a current credential for domain $D$, $q_{a,D}\geq Q_{\min}$, finite $e_{a,D}\leq e_{\max}$, and $0\leq p_a\leq p_{\max}$. The weights are nonnegative, sum to 1, and have $w_p>0$. The subscripts $q,e,\rho,p$ identify the quality, cost, reputation, and price terms. The highest score wins. Equal scores are resolved by a fixed ordering of agent identifiers. If no bid is eligible, the task remains unassigned.

\paragraph{Payment rule.} The proposed rule uses the second-highest eligible score as its payment threshold. Let $S_{(2)}$ denote that score and let $p_a$ and $S_a$ be the winning bid price and score. The threshold payment is $p_a+(p_{\max}/w_p)(S_a-S_{(2)})$. The operator assigns the task only if at least two bids are eligible and this payment does not exceed $p_{\max}$.

\section{Experimental Validation}


We compared task quality and cost under controlled model, harness, and budget settings to assess the information carried by the credential. We also quantified funding requirements for reputation manipulation under the Layer~2 attack model in Section~\ref{sec:sybil}. These evaluations address measurement scope and a specific economic threat, without validating end-to-end marketplace performance.

\subsection{Method}

\paragraph{Evaluation design.} We evaluated model and harness combinations on shared task sets. Table~\ref{tab:evaluation-design} in Appendix~\ref{app:evaluation-design} lists the comparisons and token budgets. The appendix describes the models, harnesses, and task sets. All configurations ran at temperature~0 with pinned model snapshots. Repeated runs measured variation. The earlier $\tau$-bench study found that function calling outperformed text-based ReAct on retail tasks using the same models~\citep{yao2024tau}. The experiments compare model choice, harness design, and retrieval under stated token budgets.

\paragraph{Measurements.} We report task quality, cost per solved task, and Pareto dominance under a fixed token budget. These measurements supply the Layer~1 credential fields. Over-budget tasks receive zero credit. Cost uses provider-reported usage and price snapshots for each run. Appendix~\ref{app:evaluation-methods} gives the execution, calibration, and usage-accounting details.

\paragraph{Scoring.} The GAIA comparisons use the curated task subset and official exact-match scoring~\citep{mialon2023gaia}. An LLM judge awards credit to near-misses only when it confirms equivalent meaning. We checked its agreement with human labels before its decisions affected results.

\paragraph{Statistics.} The ratio $\hat{\kappa}$ divides one configuration\textquotesingle{}s mean quality by the comparison configuration\textquotesingle{}s mean quality on the same tasks. A ratio above 1 favors the first configuration. A ratio of 1.2 means its mean quality is 20\% higher. The symbol $\kappa$ denotes the corresponding ratio of expected quality for the stated task distribution. We report the original paired-bootstrap confidence intervals and label the original approximate Wilcoxon probabilities as reported values. The comparisons with a stated decision criterion require $\hat{\kappa}\geq1.2$ with a 95\% CI excluding 1.0. A retrospective check uses within-task permutations and exact paired tests on binary outcomes. Appendix~\ref{app:paired-checks} gives the procedures and correction families. CI means confidence interval, and SD means standard deviation. A test probability is denoted by $p$.

For the grid's quality conclusions, we use the retrospective paired tests with Bonferroni correction and a significance threshold of 0.05. The omnibus family contains four tests. Exact baseline model comparisons form a family of three, and within-model harness comparisons against baseline form a separate family of 15. Failure to reject a null does not establish equivalence. The original ratio intervals and approximate probabilities are retained as reported analyses and are not simultaneous intervals across those families. Cost ratios describe a separate outcome, with bootstrap uncertainty where available. Similar solved counts do not constitute a formal equivalence test. The variance decompositions are descriptive, and no inferential claim about a model-by-harness interaction is made.

\FloatBarrier
\subsection{Results}

\subsubsection{Model capability and unrestricted retrieval}

We compared \texttt{gpt-5.6-luna} and \texttt{gpt-5.6-sol} using the same baseline harness. We then added hosted web search to each model without limiting the number of search calls. Each task had a 16,000-token budget to allow tokens for retrieval. Tasks exceeding this budget received zero credit. We also tested reflective retry in a fresh \texttt{gpt-5.6-sol} evaluation to compare its quality and cost with baseline. Reported costs include search fees.

\begin{table}[!htbp]
\centering
\begin{minipage}{0.8\textwidth}
\caption{Model and harness performance on 127 GAIA tasks at $B = 16{,}000$ tokens. Over-budget tasks receive zero credit. Success rates use all 127 tasks.}
\label{tab:followup-tests}
\small
\setlength{\tabcolsep}{4pt}
\renewcommand{\arraystretch}{1.15}
\begin{tabular*}{\linewidth}{@{\extracolsep{\fill}}llrrrr@{}}
\toprule
\tableheading{\textbf{Model}} & \tableheading{\textbf{Harness}} & \tableheading{\textbf{Solved}} & \tableheading{\textbf{Success rate}} & \tableheading{\textbf{Total tokens}} & \tableheading{\textbf{Over budget}} \\
\midrule
\texttt{gpt-5.6-luna} & Baseline & 38 & 29.9\% & 304{,}138 & 0 \\
\texttt{gpt-5.6-luna} & Web search & 48 & 37.8\% & 3{,}121{,}745 & 68 \\
\texttt{gpt-5.6-sol} & Baseline & 54 & 42.5\% & 321{,}980 & 0 \\
\texttt{gpt-5.6-sol} & Web search & 52 & 40.9\% & 3{,}184{,}671 & 70 \\
\texttt{gpt-5.6-sol} & Reflective retry & 53 & 41.7\% & 507{,}747 & 0 \\
\bottomrule
\end{tabular*}
\par\medskip
\small
\begin{tabularx}{\linewidth}{@{}>{\raggedright\arraybackslash}X>{\raggedright\arraybackslash}p{2cm}rlrl@{}}
\toprule
\tableheading{\textbf{Comparison}} & \tableheading{\textbf{Metric}} & \tableheading{\textbf{Ratio}} & \tableheading{\textbf{95\% CI}} & \tableheading{$p$} & \tableheading{\textbf{Criterion}} \\
\midrule
\texttt{gpt-5.6-sol} baseline over \texttt{gpt-5.6-luna} baseline & Success rate & 1.421 & $[1.152, 1.788]$ & 0.0045 & Met \\
\texttt{gpt-5.6-luna} web search over its baseline & Success rate & 1.263 & $[0.955, 1.710]$ & 0.148 & Not met \\
\texttt{gpt-5.6-sol} reflective retry over its baseline & Cost per solved task & 1.495 & $[1.280, 1.750]$ & N/A & Met \\
\texttt{gpt-5.6-sol} web search over its baseline & Success rate & 0.963 & $[0.745, 1.239]$ & 0.797 & N/A \\
\bottomrule
\end{tabularx}
\par\smallskip\parbox{\linewidth}{\footnotesize Each ratio divides the first configuration's metric by the second's. The criterion requires a ratio of at least 1.2 and a 95\% CI excluding 1.0. The final comparison belongs to the $2 \times 2$ model and tool analysis. It is descriptive, so the criterion does not apply. The cost comparison uses a bootstrap CI only and has no $p$ value. }
\end{minipage}
\end{table}

Table~\ref{tab:followup-tests} shows that the baseline solved 54 tasks with \texttt{gpt-5.6-sol} and 38 with \texttt{gpt-5.6-luna}, a 42.1\% increase. Model choice therefore changes the measured quality. Reflective retry on \texttt{gpt-5.6-sol} solved 53 tasks, compared with 54 for baseline. Its cost per solved task was 49.5\% higher.

Web search improved \texttt{gpt-5.6-luna}'s observed success, increasing solved tasks from 38 to 48. This is a 26.3\% increase under the 16,000-token budget. Search solved 52 tasks with \texttt{gpt-5.6-sol}, compared with 54 for baseline. The benefit of adding search therefore differed between the evaluated models.

Search exceeded the token budget on 68 tasks with \texttt{gpt-5.6-luna} and 70 with \texttt{gpt-5.6-sol}. These tasks received zero credit, but their token use and cost still counted. A credential must therefore record the tool configuration alongside the model and evaluation budget.

\begin{figure}[!htbp]
\centering
\begin{minipage}[t]{0.48\textwidth}\vspace{0pt}
\centering
\begin{tikzpicture}
\begin{axis}[
  title={(a) Token thresholds},
  width=\linewidth, height=0.78\linewidth,
  xlabel={Token budget $B'$}, ylabel={Success rate at $B'$},
  xmin=0, xmax=16800, ymin=0, ymax=0.90,
  xtick={0,4000,8000,12000,16000},
  scaled x ticks=false,
  x tick label style={/pgf/number format/fixed, /pgf/number format/1000 sep={,}},
  ymajorgrids, grid style={gray!20},
  axis line style={gray!60},
  tick label style={font=\scriptsize}, label style={font=\small},
  legend style={at={(0.03,0.97)}, anchor=north west, legend columns=1, draw=gray!40,
    fill=white, fill opacity=0.9, text opacity=1},
  legend cell align=left,
]
\addplot[vizblue, thick, mark=*, mark size=1.6pt] coordinates {
  (1000,0.1654) (2000,0.2362) (3000,0.2598) (4000,0.2835)
  (6000,0.2992) (8000,0.2992) (12000,0.2992) (16000,0.2992)};
\addlegendentry{\texttt{gpt-5.6-luna}, baseline}
\addplot[vizblue, thick, dashed, mark=square*, mark size=1.5pt] coordinates {
  (1000,0.0709) (2000,0.0787) (3000,0.0866) (4000,0.0945)
  (6000,0.1181) (8000,0.1339) (12000,0.2677) (16000,0.3780)};
\addlegendentry{\texttt{gpt-5.6-luna}, web search}
\addplot[vizverm, thick, mark=*, mark size=1.6pt] coordinates {
  (1000,0.2598) (2000,0.3228) (3000,0.3937) (4000,0.4173)
  (6000,0.4252) (8000,0.4252) (12000,0.4252) (16000,0.4252)};
\addlegendentry{\texttt{gpt-5.6-sol}, baseline}
\addplot[vizverm, thick, dashed, mark=square*, mark size=1.5pt] coordinates {
  (1000,0.0787) (2000,0.0787) (3000,0.0866) (4000,0.1024)
  (6000,0.1260) (8000,0.1339) (12000,0.2756) (16000,0.4094)};
\addlegendentry{\texttt{gpt-5.6-sol}, web search}
\addplot[vizverm, thick, densely dotted, mark=triangle*, mark size=1.9pt] coordinates {
  (1000,0.0709) (2000,0.2362) (3000,0.3150) (4000,0.3386)
  (6000,0.3780) (8000,0.4094) (12000,0.4173) (16000,0.4173)};
\addlegendentry{\texttt{gpt-5.6-sol}, reflective retry}
\end{axis}
\end{tikzpicture}
\end{minipage}\hfill
\begin{minipage}[t]{0.48\textwidth}\vspace{0pt}
\centering
\begin{tikzpicture}
\begin{axis}[
  width=\linewidth, height=0.78\linewidth,
  title={(b) Harnesses at 8,000 tokens},
  title style={font=\small}, xmin=0.5, xmax=6.5, ymin=0, ymax=0.66,
  xtick={1,...,6},
  xticklabels={baseline, ReAct, plan-exec., refl.\ retry, sys.\ heavy, notes mem.},
  x tick label style={font=\scriptsize, rotate=20, anchor=north east},
  ylabel={Solve rate}, ymajorgrids, grid style={gray!20},
  axis line style={gray!60}, tick label style={font=\scriptsize}, label style={font=\small},
  legend style={at={(0.03,0.97)}, anchor=north west, legend columns=1, draw=gray!40,
    fill=white, fill opacity=0.9, text opacity=1},
  legend cell align=left,
]
\fill[vizgreen, fill opacity=0.13] (axis cs:0.5,0.1085) rectangle (axis cs:6.5,0.1624);
\fill[vizblue,  fill opacity=0.13] (axis cs:0.5,0.2841) rectangle (axis cs:6.5,0.3331);
\fill[vizverm,  fill opacity=0.13] (axis cs:0.5,0.4272) rectangle (axis cs:6.5,0.4452);
\addplot[vizgreen, thick, mark=triangle*, mark size=2.2pt, only marks] coordinates
  {(1,0.150) (2,0.079) (3,0.118) (4,0.094) (5,0.102) (6,0.118)};
\addlegendentry{\texttt{claude-haiku-4-5}}
\addplot[vizblue, thick, mark=*, mark size=1.9pt, only marks] coordinates
  {(1,0.339) (2,0.307) (3,0.268) (4,0.299) (5,0.307) (6,0.291)};
\addlegendentry{\texttt{gpt-5.6-luna}}
\addplot[vizverm, thick, mark=square*, mark size=1.8pt, only marks] coordinates
  {(1,0.433) (2,0.441) (3,0.394) (4,0.472) (5,0.425) (6,0.362)};
\addlegendentry{\texttt{gpt-5.6-sol}}
\end{axis}
\end{tikzpicture}
\end{minipage}
\caption{Quality across budgets and harnesses. Panel (a) applies lower token thresholds to saved 16,000-token runs. Panel (b) compares harnesses at 8,000 tokens. Shaded strips show baseline mean plus or minus one SD across five runs. Model colors match across panels.}
\label{fig:dose}
\label{fig:grid}
\end{figure}
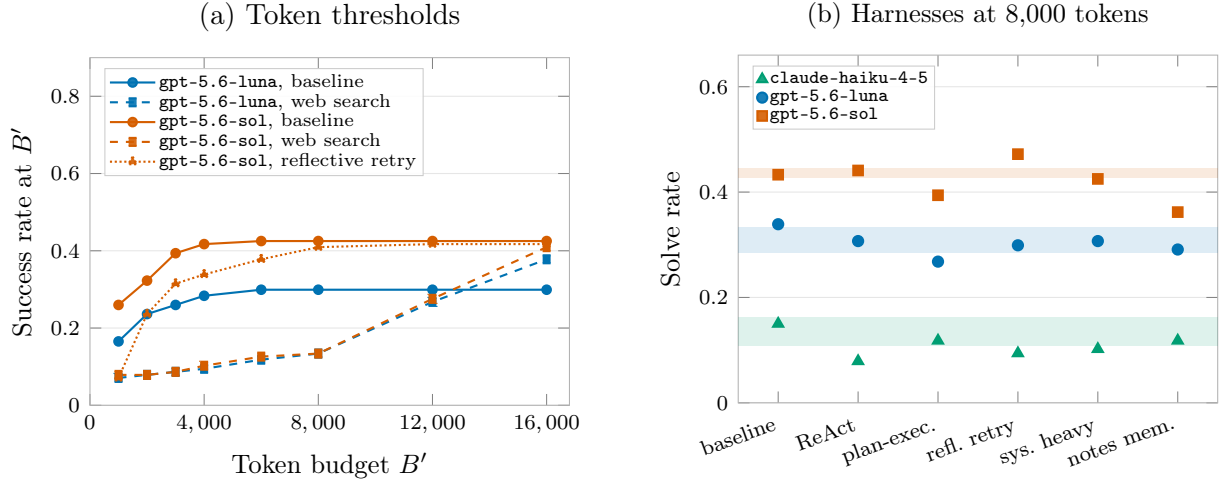

Panel (a) of Figure~\ref{fig:dose} scores completed runs again at lower token thresholds. The symbol $B'$ denotes the threshold applied to saved usage. A task keeps its final score only if its charged tokens fit within that threshold. For \texttt{gpt-5.6-luna}, web search achieves about 0.45 times baseline quality at 8,000 tokens and 1.26 times baseline quality at 16,000 tokens. Its ranking therefore changes with the evaluation budget.

\texttt{gpt-5.6-sol} without tools reaches 95\% of its full-budget quality by 4,000 tokens. Its score is 0.417 at this threshold, one quarter of the full budget, and stops increasing beyond 6,000 tokens. With search, it scores 0.409 at the full 16,000-token budget. Search scores continue improving through 16,000 tokens. These differences show why the credential records the domain's evaluation budget $B_D$.

\textbf{Reflection.} Search benefits vary with the model and token budget, supporting LEGIT credentials that bind quality measurements to the evaluated model, tools, and budget.

\subsubsection{Bounded retrieval and evaluation budget}
\label{sec:retrieval-budget}

Table~\ref{tab:retrieval} shows that search limited to two calls improved observed success for both models under the same 16,000-token budget. Solved tasks increased from 37 to 57 for \texttt{gpt-5.6-luna} and from 56 to 63 for \texttt{gpt-5.6-sol}. These are success ratios of 1.54 and 1.13. The gains came with additional search cost. For \texttt{gpt-5.6-luna}, cost per solved task was 5.7 times its no-tools baseline.

The capped runs exceeded the budget on 36 tasks for \texttt{gpt-5.6-luna} and 37 for \texttt{gpt-5.6-sol}. These counts are lower than in the separate unrestricted-search runs. Each search configuration is compared with its own no-tools baseline. The observed quality gains and added costs support recording the tool configuration, quality, cost, and budget together.

\begin{table}[!htbp]
\centering
\begin{minipage}{0.8\textwidth}
\caption{Retrieval at $B = 16{,}000$ on 127 GAIA tasks. Bounded search allows at most two tool calls per task. Baselines are fresh evaluations for this comparison. }
\label{tab:retrieval}
\small
\setlength{\tabcolsep}{4pt}
\renewcommand{\arraystretch}{1.12}
\begin{tabularx}{\linewidth}{@{}l>{\raggedright\arraybackslash}Xrrr@{}}
\toprule
\tableheading{\textbf{Model}} & \tableheading{\textbf{Harness}} & \tableheading{\textbf{Solved}} & \tableheading{\textbf{Median tokens}} & \tableheading{\textbf{Over budget}} \\
\midrule
\texttt{luna} & No tools & 37 & 2,702 & 0 \\
\texttt{luna} & Bounded search & 57 & 11,240 & 36 \\
\texttt{sol} & No tools & 56 & 2,612 & 0 \\
\texttt{sol} & Bounded search & 63 & 12,760 & 37 \\
\midrule
\multicolumn{2}{l}{\tableheading{\textbf{Search quality contrast}}} & \tableheading{$\hat{\kappa}$} & \tableheading{\textbf{95\% CI}} & \tableheading{$p$} \\
\multicolumn{2}{l}{\texttt{luna}, bounded over no tools} & 1.54 & $[1.21, 2.10]$ & 0.007 \\
\multicolumn{2}{l}{\texttt{sol}, bounded over no tools} & 1.13 & $[0.92, 1.41]$ & Not reported \\
\bottomrule
\end{tabularx}
\par\smallskip\parbox{\linewidth}{\footnotesize Bounded search costs $5.7\times$ as much per solved task on \texttt{luna}. Unrestricted search returned about 15K tokens of content per task and had median usage of about 17,000 tokens. It exceeded budget on 68 tasks and cost $17\times$ as much as its baseline per attempt. The separate capped-search run recorded 36 overruns and median usage of 11,240 tokens. }
\end{minipage}
\end{table}

Table~\ref{tab:budget-check} compares actual 4,000-token runs with estimates from saved 8,000-token runs for \texttt{gpt-5.6-luna}. The estimate counts a task only if the saved run solved it using at most 4,000 charged tokens. The direct run solved one more task for baseline, two fewer for ReAct, three more for plan-then-execute, and one more for reflective retry.

\begin{table}[!htbp]
\centering
\begin{minipage}{0.8\textwidth}
\caption{Direct evaluation at $B = 4{,}000$ against truncation from $B = 8{,}000$ for \texttt{gpt-5.6-luna}. Values are solved counts out of 127 tasks.}
\label{tab:budget-check}
\small
\setlength{\tabcolsep}{4pt}
\renewcommand{\arraystretch}{1.12}
\begin{tabularx}{\linewidth}{@{}>{\raggedright\arraybackslash}Xrrr@{}}
\toprule
\tableheading{\textbf{Harness}} & \tableheading{\textbf{Direct}} & \tableheading{\textbf{Reconstructed}} & \tableheading{\textbf{Difference}} \\
\midrule
Baseline & 40 & 39 & $+1$ \\
ReAct & 34 & 36 & $-2$ \\
Plan-then-execute & 30 & 27 & $+3$ \\
Reflective retry & 32 & 31 & $+1$ \\
\bottomrule
\end{tabularx}
\end{minipage}
\end{table}

\textbf{Reflection.} Capped search improves task success at added cost, supporting LEGIT credentials that report quality and cost together with the search limits used during evaluation.

\subsubsection{Model and harness effects}
\label{sec:factorial-results}

Table~\ref{tab:factorial} compares three models from two vendors with four harnesses on the same 127 tasks at 8,000 tokens. The baseline solved 19 tasks with \texttt{claude-haiku-4-5}, 43 with \texttt{gpt-5.6-luna}, and 55 with \texttt{gpt-5.6-sol}. Panel (b) of Figure~\ref{fig:grid} adds an elaborate system prompt and a condensed-notes memory buffer, extending the comparison to six harnesses. The tables report quality and cost for each configuration.

\begin{table}[!htbp]
\centering
\caption{Model and harness factorial grid on the shared 127-task pool at $B = 8{,}000$ tokens, with a descriptive variance decomposition on per-task scores at the right. Over-budget items are scored zero by the same-budget rule. The models are \texttt{claude-haiku-4-5}, \texttt{gpt-5.6-luna}, and \texttt{gpt-5.6-sol}. Cost per solved task is in US dollars, and the \texttt{haiku} figure uses that model's own prices.}
\label{tab:factorial}

\small
\setlength{\tabcolsep}{4pt}
\begin{minipage}[t]{0.55\textwidth}
\centering
\begin{tabular}{@{}llrrr@{}}
\toprule
\tableheading{\textbf{Model}} & \tableheading{\textbf{Harness}} & \tableheading{\textbf{Solved}} & \tableheading{\shortstack[r]{\textbf{Cost per}\\\textbf{solved task}}} & \tableheading{\shortstack[r]{\textbf{Median}\\\textbf{tokens}}} \\
\midrule
\texttt{haiku} & baseline         & 19 & 0.0143 & 425 \\
\texttt{haiku} & ReAct            & 10 & 0.0311 & 495 \\
\texttt{haiku} & plan-execute     & 15 & 0.0326 & 1{,}117 \\
\texttt{haiku} & reflective retry & 12 & 0.0706 & 2{,}191 \\
\texttt{luna}  & baseline         & 43 & 0.0086 & 2{,}256 \\
\texttt{luna}  & ReAct            & 39 & 0.0096 & 2{,}220 \\
\texttt{luna}  & plan-execute     & 34 & 0.0113 & 2{,}677 \\
\texttt{luna}  & reflective retry & 38 & 0.0140 & 3{,}775 \\
\texttt{sol}   & baseline         & 55 & 0.1206 & 2{,}704 \\
\texttt{sol}   & ReAct            & 56 & 0.1090 & 2{,}398 \\
\texttt{sol}   & plan-execute     & 50 & 0.1562 & 3{,}230 \\
\texttt{sol}   & reflective retry & 60 & 0.1469 & 3{,}627 \\
\bottomrule
\end{tabular}
\end{minipage}
\hfill
\begin{minipage}[t]{0.38\textwidth}
\centering
\begin{tabular}{@{}lrrrr@{}}
\toprule
\tableheading{\textbf{Source}} & \tableheading{\textbf{SS}} & \tableheading{\textbf{df}} & \tableheading{$F$} & \tableheading{$\eta^2$} \\
\midrule
Model       & 27.11  & 2    & 72.96 & 0.088 \\
Harness     & 0.46   & 3    & 0.82  & 0.002 \\
Interaction & 0.63   & 6    & 0.56  & 0.002 \\
Residual    & 280.91 & 1512 &       & 0.909 \\
\bottomrule
\end{tabular}
\end{minipage}
\par\smallskip\parbox{\textwidth}{\footnotesize SS is the sum of squared deviations assigned to each source. The degrees of freedom are df. The statistic $F$ divides a source\textquotesingle s mean square by the residual mean square. Paired tests appear in Table~\ref{tab:paired-checks}. The effect size $\eta^2$ is its share of total SS. This statistical symbol is separate from the reputation decay factor $\eta$.}
\end{table}

\begin{table}[!htbp]
\centering
\begin{minipage}{0.8\textwidth}
\caption{Quality contrasts and the extended six-harness analysis on GAIA at $B = 8{,}000$. Model contrasts use the baseline harness.}
\label{tab:quality-contrasts}
\small
\setlength{\tabcolsep}{4pt}
\renewcommand{\arraystretch}{1.12}
\begin{tabularx}{\linewidth}{@{}>{\raggedright\arraybackslash}Xrr@{}}
\toprule
\tableheading{\textbf{Contrast}} & \tableheading{$\hat{\kappa}$} & \tableheading{\textbf{95\% CI}} \\
\midrule
\texttt{sol} over \texttt{luna} & 1.28 & $[1.09, 1.55]$ \\
\texttt{luna} over \texttt{haiku} & 2.26 & $[1.64, 3.55]$ \\
\texttt{sol} over \texttt{haiku} & 2.90 & $[2.04, 4.69]$ \\
\texttt{sol} notes over baseline & 0.84 & $[0.70, 0.96]$ \\
\midrule
\tableheading{\textbf{Six-harness analysis}} & \tableheading{$F$} & \tableheading{\textbf{Result}} \\
Model & 101.4 & Descriptive \\
Harness & 0.71 & Descriptive \\
Interaction & 0.54 & Descriptive \\
\bottomrule
\end{tabularx}
\par\smallskip\parbox{\linewidth}{\footnotesize The notes harness solves 46 tasks on \texttt{sol}, against 55 for baseline. }
\end{minipage}
\end{table}

Baseline success is higher for \texttt{gpt-5.6-sol} than \texttt{gpt-5.6-luna}, and higher for \texttt{gpt-5.6-luna} than \texttt{claude-haiku-4-5}, as shown in Table~\ref{tab:quality-contrasts}. All three exact paired baseline model comparisons remain significant after correction. The omnibus model tests have adjusted $p=0.00020$ for both the four-harness and six-harness grids. The corresponding harness tests have adjusted $p=0.15100$ and $p=0.15280$, and none of the 15 exact within-model harness comparisons remains significant. Tables~\ref{tab:paired-checks} and~\ref{tab:exact-pairs} give the full results.

The observed harness differences are therefore descriptive. On \texttt{gpt-5.6-sol}, ReAct solved 56 tasks and reflective retry solved 60, compared with 55 for baseline. The notes harness solved 46. Baseline solved the most tasks among the tested harnesses for \texttt{gpt-5.6-luna} and \texttt{claude-haiku-4-5}. These counts do not establish an overall harness-quality effect or a general benefit from additional orchestration. The reported unadjusted interval for the notes contrast in Table~\ref{tab:quality-contrasts} should be read alongside its nonsignificant corrected paired test.

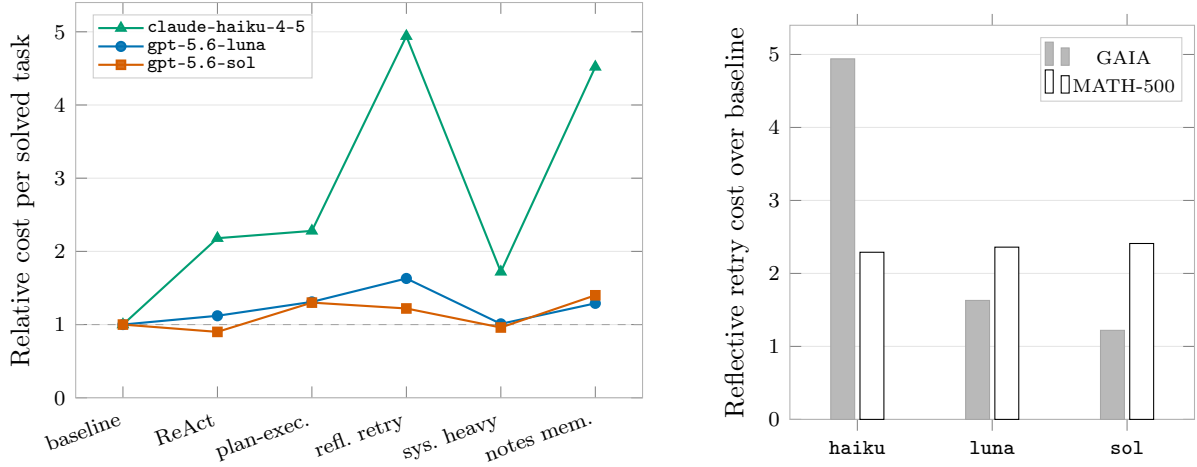
\begin{figure}[!htbp]
\centering
\begin{minipage}{0.55\textwidth}
\centering
\begin{tikzpicture}
\begin{axis}[
  width=\textwidth, height=0.75\textwidth,
  xmin=0.5, xmax=6.5, ymin=0, ymax=5.4,
  xtick={1,...,6},
  xticklabels={baseline, ReAct, plan-exec., refl.\ retry, sys.\ heavy, notes mem.},
  x tick label style={font=\scriptsize, rotate=20, anchor=north east},
  ylabel={Relative cost per solved task},
  ymajorgrids, grid style={gray!20}, axis line style={gray!60},
  tick label style={font=\scriptsize}, label style={font=\small},
  legend style={at={(0.03,0.97)}, anchor=north west, draw=gray!40,
    fill=white, fill opacity=0.9, text opacity=1},
  legend cell align=left,
]
\addplot[gray!70, dashed, domain=0.5:6.5, forget plot] {1};
\addplot[vizgreen, thick, mark=triangle*, mark size=2.0pt] coordinates
  {(1,1.00) (2,2.18) (3,2.28) (4,4.94) (5,1.72) (6,4.52)};
\addlegendentry{\texttt{claude-haiku-4-5}}
\addplot[vizblue, thick, mark=*, mark size=1.7pt] coordinates
  {(1,1.00) (2,1.12) (3,1.31) (4,1.63) (5,1.01) (6,1.29)};
\addlegendentry{\texttt{gpt-5.6-luna}}
\addplot[vizverm, thick, mark=square*, mark size=1.6pt] coordinates
  {(1,1.00) (2,0.90) (3,1.30) (4,1.22) (5,0.96) (6,1.40)};
\addlegendentry{\texttt{gpt-5.6-sol}}
\end{axis}
\end{tikzpicture}
\end{minipage}\hfill
\begin{minipage}{0.42\textwidth}
\centering
\begin{tikzpicture}
\begin{axis}[
  width=\textwidth, height=0.98\textwidth,
  ybar, bar width=9pt,
  xmin=0.5, xmax=3.5, ymin=0, ymax=5.4,
  xtick={1,2,3},
  xticklabels={\texttt{haiku}, \texttt{luna}, \texttt{sol}},
  x tick label style={font=\scriptsize},
  ylabel={Reflective retry cost over baseline},
  ymajorgrids, grid style={gray!20}, axis line style={gray!60},
  tick label style={font=\scriptsize}, label style={font=\small},
  legend style={at={(0.97,0.97)}, anchor=north east, draw=gray!40},
]
\addplot[fill=gray!55, draw=gray!70] coordinates {(1,4.94) (2,1.63) (3,1.22)};
\addlegendentry{GAIA}
\addplot[fill=white, draw=black] coordinates {(1,2.29) (2,2.36) (3,2.41)};
\addlegendentry{MATH-500}
\end{axis}
\end{tikzpicture}
\end{minipage}
\caption{The left panel compares each harness's cost per solved task with the same model's baseline on GAIA. Model colors match Figure~\ref{fig:grid}, right panel. The dashed line marks equal cost. The right panel shows reflective retry's extra cost on both task domains. Extra cost occurs everywhere and is largest on the weakest model.}
\label{fig:efficiency}
\end{figure}

\begin{figure}[!htbp]
\centering
\begin{tikzpicture}
\begin{axis}[
  width=0.64\textwidth, height=0.40\textwidth,
  xmode=log, log ticks with fixed point,
  xmin=0.006, xmax=0.24, ymin=0.05, ymax=0.52,
  xlabel={Cost per solved task in \$, log scale}, ylabel={Solve rate},
  ymajorgrids, grid style={gray!20}, axis line style={gray!60},
  tick label style={font=\small}, label style={font=\small},
  legend style={at={(0.97,0.05)}, anchor=south east, draw=gray!40,
    fill=white, fill opacity=0.9, text opacity=1},
  legend cell align=left,
]
\addplot[gray!70, dashed, thick] coordinates
  {(0.0086,0.339) (0.1090,0.441) (0.1469,0.472)};
\addlegendentry{Pareto frontier}
\addplot[vizgreen, only marks, mark=triangle*, mark size=2.4pt] coordinates
  {(0.0143,0.150) (0.0311,0.079) (0.0326,0.118) (0.0706,0.094) (0.0246,0.102) (0.0647,0.118)};
\addlegendentry{\texttt{claude-haiku-4-5}}
\addplot[vizblue, only marks, mark=*, mark size=2.0pt] coordinates
  {(0.0086,0.339) (0.0096,0.307) (0.0113,0.268) (0.0140,0.299) (0.0087,0.307) (0.0111,0.291)};
\addlegendentry{\texttt{gpt-5.6-luna}}
\addplot[vizverm, only marks, mark=square*, mark size=1.9pt] coordinates
  {(0.1206,0.433) (0.1090,0.441) (0.1562,0.394) (0.1469,0.472) (0.1161,0.425) (0.1694,0.362)};
\addlegendentry{\texttt{gpt-5.6-sol}}
\node[font=\scriptsize, anchor=south west] at (axis cs:0.0088,0.345) {baseline};
\node[font=\scriptsize, anchor=south east] at (axis cs:0.1090,0.447) {ReAct};
\node[font=\scriptsize, anchor=south west] at (axis cs:0.1469,0.476) {refl.\ retry};
\end{axis}
\end{tikzpicture}
\caption{Quality and cost per solved task for all eighteen configurations at $B = 8{,}000$. The dashed line joins the Pareto frontier. It contains \texttt{luna} baseline, \texttt{sol} ReAct, and \texttt{sol} reflective retry at very different trade-offs. Every other configuration is dominated. A quality-only ranking would hide the entire horizontal axis.}
\label{fig:pareto}
\end{figure}
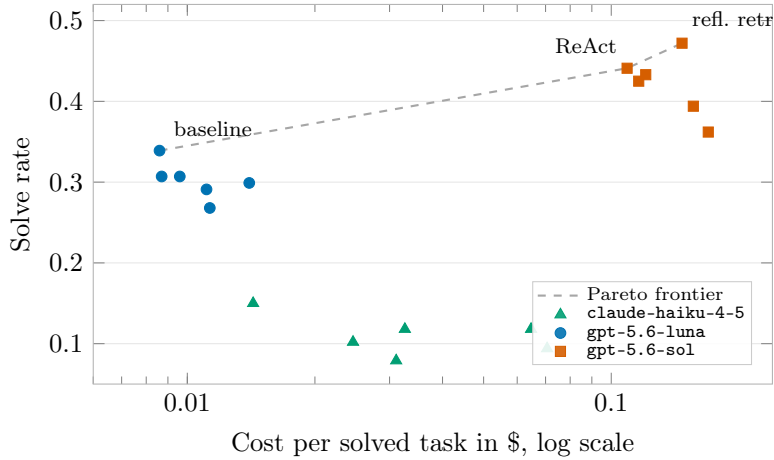

Figure~\ref{fig:efficiency} shows cost differences within each model. The most expensive harness costs 1.43 times the least expensive on \texttt{gpt-5.6-sol}, 1.63 times on \texttt{gpt-5.6-luna}, and 4.95 times on \texttt{claude-haiku-4-5}. On \texttt{claude-haiku-4-5}, reflective retry uses about five times the baseline tokens and solves fewer tasks. The cost spread is largest for the model with the lowest success in these runs. Figure~\ref{fig:pareto} places three of the eighteen configurations on the Pareto frontier, meaning no other tested configuration achieves at least as much quality at no greater cost, with an improvement in either measure. A quality-only ranking would hide these cost differences.

\begin{figure}[!htbp]
\centering
\begin{tikzpicture}
\input{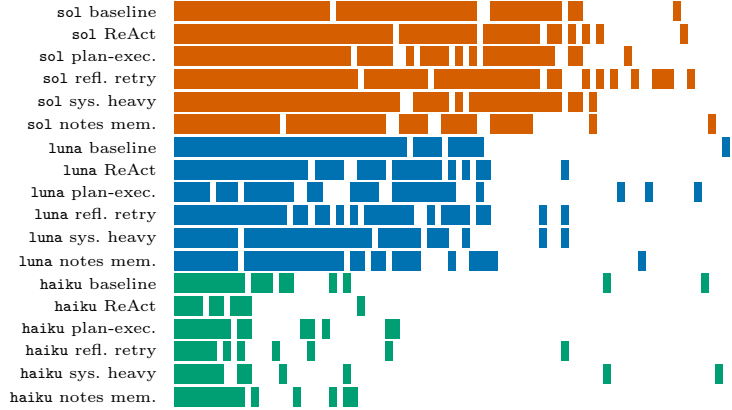}
\end{tikzpicture}
\caption{Task outcomes underlying the grid statistics. Each row represents a configuration. Each cell represents a task in the shared pool and is filled when that configuration solves it under the same-budget rule. The left side contains easy tasks shared across configurations. The middle band contains tasks solved by some models but not others. The right side contains unsolved tasks. Rows differ far less within a model than across models.}
\label{fig:matrix}
\end{figure}

Figure~\ref{fig:matrix} shows which tasks each configuration solved. The middle band contains tasks solved by some models but not others. Changing the harness within a model changes fewer outcomes. This task-level view helps explain why model differences are more visible than harness differences in the aggregate results.

\textbf{Reflection.} The grid supports model-quality differences and documents cost differences among harness configurations. Corrected harness-quality tests are inconclusive. Recording both quality and cost preserves these distinctions without implying that more elaborate harnesses improve task success.

\subsubsection{Repeatability, scoring, and domain transfer}
\label{sec:validation-transfer}

Table~\ref{tab:repeatability} reports five baseline runs per model. The standard deviation in solved counts was 3.4 tasks for \texttt{claude-haiku-4-5}, 3.1 for \texttt{gpt-5.6-luna}, and 1.1 for \texttt{gpt-5.6-sol}. Relative to each model's mean, these are roughly 20\%, 8\%, and 2\%. Runs varied even at temperature~0, consistent with~\citet{ouyang2023nondeterminism}.

\begin{table}[!htbp]
\centering
\begin{minipage}[t]{0.49\textwidth}
\centering
\caption{Baseline repeatability on 127 GAIA tasks at $B = 8{,}000$ and temperature~0, with the invocation cache disabled.}
\label{tab:repeatability}

\footnotesize
\setlength{\tabcolsep}{3pt}
\renewcommand{\arraystretch}{1.12}
\begin{tabularx}{\linewidth}{@{}>{\raggedright\arraybackslash}Xrrr@{}}
\toprule
\tableheading{\textbf{Model}} & \tableheading{\textbf{Runs}} & \tableheading{\textbf{SD in tasks}} & \tableheading{\textbf{SD rel.\ to mean}} \\
\midrule
\texttt{haiku} & 5 & 3.4 & roughly 20\% \\
\texttt{luna} & 5 & 3.1 & roughly 8\% \\
\texttt{sol} & 5 & 1.1 & roughly 2\% \\
\bottomrule
\end{tabularx}
\par\smallskip\parbox{\linewidth}{\footnotesize The \texttt{luna} runs solved 43, 37, 42, 36, and 38 tasks, with mean 39.2. }
\end{minipage}\hfill
\begin{minipage}[t]{0.49\textwidth}
\centering
\caption{MATH-500 transfer on 127 tasks at the same $B = 8{,}000$ budget. Cost ratios compare reflective retry with the same model's baseline.}
\label{tab:math-transfer}

\footnotesize
\setlength{\tabcolsep}{3pt}
\renewcommand{\arraystretch}{1.12}
\begin{tabularx}{\linewidth}{@{}>{\raggedright\arraybackslash}Xrrr@{}}
\toprule
\tableheading{\textbf{Model}} & \tableheading{\textbf{Baseline solved}} & \tableheading{\textbf{Retry solved}} & \tableheading{\textbf{Cost ratio}} \\
\midrule
\texttt{haiku} & 118 & 120 & $2.29\times$ \\
\texttt{luna} & 123 & 125 & $2.36\times$ \\
\texttt{sol} & 127 & 127 & $2.41\times$ \\
\bottomrule
\end{tabularx}
\par\smallskip\parbox{\linewidth}{\footnotesize The baseline model contrast between \texttt{sol} and \texttt{haiku} gives $\hat{\kappa} = 1.08$ with 95\% CI $[1.03, 1.13]$. Cost ratios are the reported ratios in Figure~\ref{fig:efficiency}.}
\end{minipage}
\end{table}

Judgments also agree across model families. Reassessment of 1,612 cached judgments by \texttt{claude-haiku-4-5} agreed with \texttt{gpt-5.6-luna} on 99.1\% under identical instructions. Cohen's kappa was 0.94.

Table~\ref{tab:math-transfer} compares baseline and reflective retry on 127 MATH-500 tasks converted to the evaluation system's task format. Retry increased solved counts from 118 to 120 for \texttt{claude-haiku-4-5} and from 123 to 125 for \texttt{gpt-5.6-luna}. Both \texttt{gpt-5.6-sol} configurations solved all 127 tasks. Retry cost 2.29, 2.36, and 2.41 times as much per solved task, respectively. The cost difference therefore persists on mathematics tasks.

\textbf{Reflection.} Run variation, scoring checks, and cost differences on mathematics support LEGIT credentials that retain evaluation evidence and report quality and cost for each domain.

\FloatBarrier
\subsection{Measured Credential Profiles}
\label{sec:credential-profiles}

The left panel of Figure~\ref{fig:hex} combines the measured quality, cost, and budget results into agent profiles. Among these configurations, the \texttt{gpt-5.6-luna} baseline is most efficient on GAIA and MATH-500. The \texttt{gpt-5.6-sol} configurations lead on task quality. The profiles show why a credential needs both quality and cost.

The right panel of Figure~\ref{fig:card} is an example display of a structured credential using measured \texttt{gpt-5.6-luna} baseline results. The quality, cost per solved task, and budget values come from the evaluations. The shortened signature, expiry date, and verification mark illustrate presentation fields. An actual verification decision requires the signed record, a trusted issuer key, and the applicable scope, age, and status checks.

The label \texttt{haiku} denotes \texttt{claude-haiku-4-5}. The label \texttt{luna} denotes \texttt{gpt-5.6-luna}. The label \texttt{sol} denotes \texttt{gpt-5.6-sol}. Clockwise from the top, the profile axes show quality on GAIA level 1, GAIA level 2 and higher, and MATH-500. The next axes show efficiency on GAIA and MATH-500, meaning inverse cost per solved task. The final axis shows budget sensitivity, measured as the share of full-budget quality retained when saved GAIA runs are scored at half the evaluation budget. It is a threshold reconstruction, not a direct measurement of an agent adapting to that lower budget. Each axis is normalized to its best agent. The card uses $q$ for domain quality, $\mathsf{CoP}$ for cost per solved task, and $B$ for the token budget. It displays evidence assurance, expiry, and issuer-signature information from the underlying record.

\textbf{Reflection.} The structured profile retains the measurements and conditions needed for comparisons. Its visual presentation is optional and introduces no additional assurance.

\FloatBarrier
\subsection{Sybil Attack Economics}
\label{sec:sybil}\label{sec:threats}

We assess the economic barrier to an attacker accumulating misleading reputation across multiple controlled identities. Each Sybil identity starts with a uniform reputation prior and $\rho_0=0.5$. The attacker locks $s_{\min}$ per identity and spends a fee of at least $f_{\min}$ per credited interaction. Perfect feedback gives the best-case interaction count $n_{\mathrm{rep}}$ needed to reach $\rho_{\mathrm{target}}$. The model assumes every identity incurs its own deposit and fees, deposits stay locked throughout accumulation, and fees cannot be recovered through self-dealing. It conditions on the specified feedback stream without establishing how an attacker obtains accepted positive reports. Appendix~\ref{app:security} gives the closed-form count and its agreement with direct iteration.

\begin{table}[!htbp]
\centering
\caption{Capital and fees required to accumulate reputation for a Sybil identity. The left panel gives the perfect-feedback interactions $n_{\text{rep}}$ needed to raise a fresh identity from $\rho_0 = 0.5$ to $\rho_{\text{target}}$, from direct iteration of the discounted Beta update. The right panel gives deposit plus spent fees per identity at $\eta=0.9$ in USD. The sum is $s_{\min}+n_{\mathrm{rep}}f_{\min}$. The dollar-valued column headings specify the locked deposit. }
\label{tab:sybil}

\small
\setlength{\tabcolsep}{4pt}
\renewcommand{\arraystretch}{1.12}
\begin{minipage}[t]{0.36\textwidth}
\centering
\begin{tabular}{@{}lrrr@{}}
\toprule
\tableheading{$\rho_{\text{target}}$} & \tableheading{$\eta = 0.80$} & \tableheading{$\eta = 0.90$} & \tableheading{$\eta = 0.95$} \\
\midrule
0.70 & 2  & 2  & 2  \\
0.80 & 3  & 3  & 3  \\
0.90 & 5  & 6  & 7  \\
0.95 & 7  & 10 & 13 \\
0.99 & 14 & 23 & 35 \\
\bottomrule
\end{tabular}
\end{minipage}
\hfill
\begin{minipage}[t]{0.57\textwidth}
\centering
\begin{tabular}{@{}rrr rrrr@{}}
\toprule
\tableheading{$f_{\min}$} & \tableheading{$\rho_{\text{target}}$} & \tableheading{$n_{\text{rep}}$} & \tableheading{\$10} & \tableheading{\$50} & \tableheading{\$100} & \tableheading{\$500} \\
\midrule
\$0.10 & 0.90 & 6 & \$10.60 & \$50.60 & \$100.60 & \$500.60 \\
\$0.10 & 0.95 & 10 & \$11.00 & \$51.00 & \$101.00 & \$501.00 \\
\$0.10 & 0.99 & 23 & \$12.30 & \$52.30 & \$102.30 & \$502.30 \\
\$1.00 & 0.90 & 6 & \$16.00 & \$56.00 & \$106.00 & \$506.00 \\
\$1.00 & 0.95 & 10 & \$20.00 & \$60.00 & \$110.00 & \$510.00 \\
\$1.00 & 0.99 & 23 & \$33.00 & \$73.00 & \$123.00 & \$523.00 \\
\bottomrule
\end{tabular}
\end{minipage}
\end{table}

Table~\ref{tab:sybil} reports capital plus fees needed to fund the modeled attack. Fees range from \$0.60 to \$23 per identity. The deposit is the larger component for some, but not all, displayed settings.

\begin{figure}[!htbp]
\centering
\begin{tikzpicture}
\begin{axis}[
  width=0.68\textwidth, height=0.46\textwidth,
  ymode=log, log ticks with fixed point,
  xmin=0.68, xmax=1.005, ymin=1.5, ymax=4200,
  xtick={0.7,0.8,0.9,0.95,0.99},
  x tick label style={font=\small}, xlabel={Target reputation $\rho_{\text{target}}$},
  ylabel={Interactions $n_{\text{rep}}$},
  ymajorgrids, grid style={gray!20}, axis line style={gray!60},
  tick label style={font=\small}, label style={font=\small},
  legend style={at={(0.03,0.97)}, anchor=north west, draw=gray!40,
    fill=white, fill opacity=0.9, text opacity=1},
  legend cell align=left,
]
\addplot[gray!60, thick, mark=*, mark size=1.5pt] coordinates
  {(0.7,2) (0.8,3) (0.9,5) (0.95,7) (0.99,14)};
\addlegendentry{perfect feedback, $\eta=0.80$}
\addplot[gray!90, thick, mark=*, mark size=1.5pt] coordinates
  {(0.7,2) (0.8,3) (0.9,6) (0.95,10) (0.99,23)};
\addlegendentry{perfect feedback, $\eta=0.90$}
\addplot[black, thick, mark=*, mark size=1.5pt] coordinates
  {(0.7,2) (0.8,3) (0.9,7) (0.95,13) (0.99,35)};
\addlegendentry{perfect feedback, $\eta=0.95$}
\addplot[vizverm, thick, dashed, mark=square*, mark size=1.6pt] coordinates
  {(0.7,2.7) (0.8,4.7) (0.9,12.4) (0.95,29.9) (0.99,165.8)};
\addlegendentry{90\% success, $\eta=0.90$}
\addplot[vizverm, thick, densely dotted, mark=triangle*, mark size=2.0pt] coordinates
  {(0.7,4.0) (0.8,8.7) (0.9,37.1) (0.95,158) (0.99,2073)};
\addlegendentry{80\% success, $\eta=0.90$}
\node[font=\scriptsize, vizverm, anchor=east, align=right] at (axis cs:0.985,2073)
  {not reached within 5,000\\interactions in 34.1\% of trials};
\end{axis}
\end{tikzpicture}
\caption{Interactions needed to reach a reputation target. Solid curves use perfect feedback. Noisy curves use independent binary feedback with success probabilities 0.9 and 0.8 at $\eta=0.9$. Each point is the mean among trials reaching the target within 5,000 interactions, using 1,000 trials and seeds 0 through 999. At 80\% success, 341 trials do not reach 0.99 within the horizon.}
\label{fig:sybil}
\end{figure}
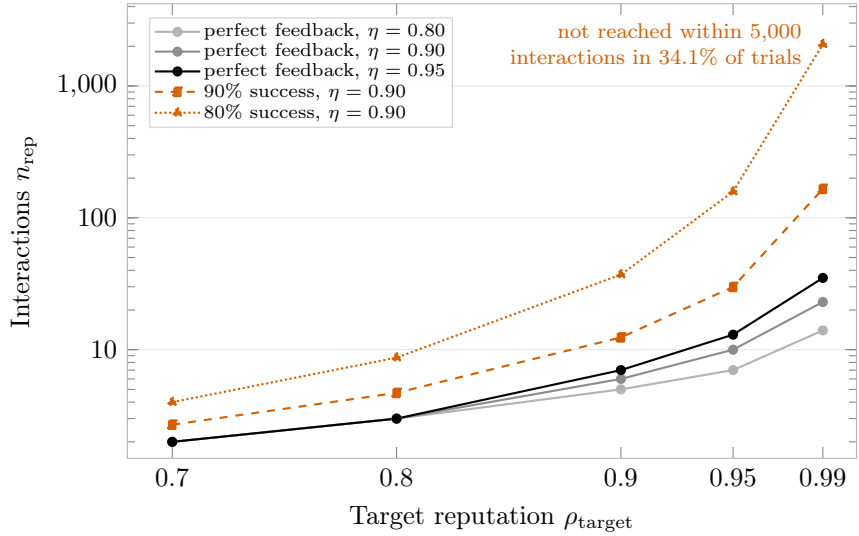

Figure~\ref{fig:sybil} shows the sensitivity to noisy feedback. At success probability 0.9, reaching reputation 0.99 takes about 166 interactions, against 23 with perfect feedback. At probability 0.8, 34.1\% of trials do not reach the target within 5,000 interactions. The plotted mean of 2,073 uses only the 659 trials that reach it.

\textbf{Reflection.} Deposits and nonrecoverable fees impose funding requirements under this model. Locked capital can be returned and is distinct from spent fees. The calculation does not prove attack unprofitability, detect common identity ownership, or establish resistance to forged feedback, coordinated manipulation, and other attacks. It provides a bounded assessment of reputation-manipulation costs for comparing protocol settings.

\FloatBarrier
\section{Conclusion}


\framework{} connects agent certification, reputation records, and proposed marketplace allocation through signed, structured credentials. Each credential binds quality and cost to configuration, domain, budget, evaluation evidence, and temporal scope. Charts and cards are optional views. Controlled comparisons support model-quality differences and document within-model cost differences and budget-sensitive retrieval rankings. Corrected harness-quality tests remain inconclusive. These results motivate the credential fields. The Sybil analysis quantifies locked capital and spent fees under explicit feedback and funding assumptions.


\bibliographystyle{unsrtnat}
\bibliography{references}

\appendix

\section{Sybil Capital and Fee Requirements}
\label{app:security}

An attacker must lock at least $N_{\mathrm{Sybil}}s_{\min}$ in deposits. Building reputation also spends at least $N_{\mathrm{Sybil}}n_{\mathrm{rep}}f_{\min}$ in non-recoverable fees. The sum is a lower bound on the funding required while all deposits remain locked. The analysis assumes each Sybil identity pays its own fees and that the attacker cannot recycle deposits during this period.

For perfect feedback and a uniform prior, the update has a closed form. For $0<\eta<1$ and interaction count $n$,
\begin{equation}
\alpha^{(n)}=\eta^n+\frac{1-\eta^n}{1-\eta},\qquad
\beta^{(n)}=\eta^n,\qquad
\rho^{(n)}=\frac{1-\eta^{n+1}}{1+(1-2\eta)\eta^n}.
\end{equation}
Let $r=\rho_{\mathrm{target}}$ with $1/2<r<1$. The required count is
\begin{equation}
n_{\mathrm{rep}}=
\left\lceil\frac{\log\!\left((1-r)/(\eta+r(1-2\eta))\right)}{\log\eta}\right\rceil.
\end{equation}
For $\eta=1$, the score is $(n+1)/(n+2)$ and
$n_{\mathrm{rep}}=\lceil(2r-1)/(1-r)\rceil$.
Targets at or below $1/2$ require no interactions. We checked these expressions against direct iteration for 36 combinations of longevity factor and target. All counts agreed.

The noisy-feedback check uses independent Bernoulli outcomes with success probabilities 0.9 and 0.8. Each target uses 1,000 trials with seeds 0 through 999 and a 5,000-interaction horizon. Means in Figure~\ref{fig:sybil} exclude trials that do not reach the target. At success probability 0.8 and target 0.99, 659 trials reach the target and 341 do not. The mean among reaching trials is 2,072.95 interactions.

\section{Evaluation Design}
\label{app:evaluation-design}

\begin{table}[!htbp]
\centering
\begin{minipage}{0.8\textwidth}
\caption{Agent evaluation design. Each benchmark comparison uses 127 shared tasks unless noted otherwise. The six-harness comparison adds an elaborate system prompt and condensed-notes memory.}
\label{tab:evaluation-design}
\small
\setlength{\tabcolsep}{4pt}
\renewcommand{\arraystretch}{1.12}
\begin{tabularx}{\linewidth}{@{}lXr@{}}
\toprule
\tableheading{\textbf{Comparison}} & \tableheading{\textbf{Scope}} & \tableheading{\textbf{Token budget}} \\
\midrule
Calibration & 2 configurations on 6 tasks & 8,000 \\
Model and tool diagnostics & 2 models across 5 configurations & 16,000 \\
Factorial comparison & 3 models with 4 harnesses & 8,000 \\
Extended harness comparison & 3 models with 6 harnesses & 8,000 \\
Baseline repeatability & 5 runs per model with invocation cache disabled & 8,000 \\
Budget reconstruction check & 4 \texttt{luna} harnesses, compared with truncation from 8,000 tokens & 4,000 \\
Bounded retrieval & \texttt{luna} and \texttt{sol}, at most 2 searches per task & 16,000 \\
Mathematics transfer & Baseline and reflective retry on MATH-500 & 8,000 \\
\bottomrule
\end{tabularx}
\end{minipage}
\end{table}

\paragraph{Harnesses.} The registry contains six configurations. The baseline makes one model call. ReAct runs a thought-and-action loop for up to four turns. Plan-then-execute makes a planning call followed by an execution call. Reflective retry answers, critiques, and revises in three calls. The elaborate-prompt harness makes one call with a longer system prompt. The notes-memory harness extracts notes, extends them, and answers from them in three calls. Each harness checks estimated input usage and caps output tokens. Estimates can overshoot the allowance, and hosted retrieval can consume additional server-side tokens. Cache reads are exempt from the token threshold but included in monetary cost.

\paragraph{Models.} OpenAI provides the evaluated \texttt{gpt-5.6-luna} and \texttt{gpt-5.6-sol} models. Anthropic provides \texttt{claude-haiku-4-5}. Short labels in tables and plots are \texttt{luna}, \texttt{sol}, and \texttt{haiku}. Each run records its input, output, and cache-read prices. The cost figures use those run-specific snapshots.

\paragraph{Task pools.} The GAIA comparisons draw from the text-only subset of the benchmark. Records with an empty or unknown reference answer are dropped, and so is every record carrying an attached file. All GAIA levels remain in the pool, and every comparison requests the reasoning axis. Task selection walks a portable sha256 index from the run seed, so a shared seed gives every configuration an identical 127-task set and licenses paired statistics. The mathematics domain converts the MATH-500 test split into the same task format and draws 127 tasks from it under a separate seed. Conversion keeps the original problem text, appends one instruction about the form of the answer, and stores the reference answer as a string.

\paragraph{Calibration.} The pilot ran two configurations on six tasks at $B=8{,}000$ to target a 40--60\% solve rate. Its median solve rate was 50\%. Binary outcomes have greatest variance near solve probability 0.5, which motivated the calibration range. The pilot exposed a default timeout that failed slow multi-call runs. Later comparisons used a 360-second timeout. Judge calibration against human-labeled traces preceded the evaluations.

\paragraph{Model and tool diagnostics.} The diagnostics block ran five configurations across two models over 127 GAIA tasks at $B = 16{,}000$. The configurations are \texttt{gpt-5.6-luna} baseline, \texttt{gpt-5.6-luna} with hosted web search, \texttt{gpt-5.6-sol} baseline, \texttt{gpt-5.6-sol} with hosted web search, and \texttt{gpt-5.6-sol} reflective retry. Hosted search ran without a call cap in this block, which is why retrieval consumed the budget.

\paragraph{Factorial comparison.} The factorial block crossed three models with four harnesses over 127 GAIA tasks at $B = 8{,}000$ under one shared seed. Each of the twelve configurations ran once. The in-run cap admits a small overshoot on the input side, so scoring over-budget items as failures after execution is what enforces the same-budget rule.

\paragraph{Extended harness comparison.} The six-harness comparison keeps the three models, the task seed, and the budget of the factorial block, and adds two harnesses. The \texttt{system\_heavy} harness supplies an elaborate system prompt. The \texttt{notes\_memory} harness supplies a condensed-notes buffer. Both ran in a later round with the invocation cache disabled, while the four-harness grid ran with the cache enabled.

\paragraph{Baseline repeatability.} The baseline harness repeated five times per model at $B = 8{,}000$ on the same task seed. The invocation cache was disabled for these runs, so each replicate re-invokes the model instead of replaying a stored output. The \texttt{gpt-5.6-luna} replicates solved 43, 37, 42, 36, and 38 tasks.

\paragraph{Budget reconstruction check.} The four tool-less harnesses on \texttt{gpt-5.6-luna} ran directly at $B = 4{,}000$ on the same task seed as the $B = 8{,}000$ grid. The truncation estimate counts an item at the lower budget only when the $B = 8{,}000$ run both solved that item and charged no more than 4,000 tokens.

\paragraph{Bounded retrieval.} The retrieval block ran a search-equipped baseline on two models against a fresh tool-less baseline, over 127 GAIA tasks at $B = 16{,}000$. The cap allows the hosted search tool at most two calls per task, and each search call carries a fixed price folded into reported cost. This block uses its own sampling seed, so its baselines pair with its capped-search configurations and not with the $B = 8{,}000$ grid.

\paragraph{Mathematics transfer.} The baseline and reflective-retry harnesses ran on all three models over 127 converted MATH-500 tasks at $B = 8{,}000$. This block uses a separate seed and a separate evaluation database from the GAIA comparisons.

\section{Evaluation Details}
\label{app:evaluation-methods}

We checked the pipeline against deterministic mock services at no cost before calibrating the budget.

\paragraph{Experimental system.} The system runs model and harness combinations on task sets selected by a shared seed. It records outputs, quality, and provider-reported usage. Input, output, and cache-read prices determine monetary cost. For $\ell$ tasks per domain at common allowance $B$, the nominal charged-token allocation is $|\mathrm{configurations}|\,|\mathcal{D}|\,\ell B$. Here $\ell$ counts tasks per domain. Scoring assigns over-budget tasks zero credit, while all recorded usage still contributes to cost.

\paragraph{Credential records.} The evaluation system signs credential manifests with Ed25519 and provides a policy-based verification API.

\paragraph{Statistics.} The original ratio intervals use 2,000 paired percentile-bootstrap resamples of task indices. A resample with zero denominator is omitted. The retained approximate Wilcoxon probabilities are reported values. The retrospective tests below use exact binary comparisons and explicit correction families.

\paragraph{Usage and scoring.} Token counts are agent-reported, meaning harnesses we control relay provider usage. An inexpensive model served as the LLM judge. We measured agreement with human labels before the judge's decisions affected results.

\section{Paired Statistical Checks}
\label{app:paired-checks}

The retrospective checks use the binary task-outcome matrix underlying Figure~\ref{fig:matrix}. Each of its 127 columns is a shared task. Row totals reproduce all 18 reported solved counts. Archived task-level records also confirm exclusive win and loss counts for 45 comparisons among ten configurations.

For an omnibus model check, each task's score is averaged across harnesses. For a harness check, it is averaged across models. Within each task we permute the tested labels and recompute the sum of squared deviations of the marginal means from their grand mean, multiplied by 127. The null assumes the tested labels are exchangeable within a task. Each test uses 19,999 random permutations and seed 20260907. The probability estimate adds one to both the exceedance count and permutation count. Bonferroni correction covers the four omnibus tests shown in Table~\ref{tab:paired-checks}. These checks were performed during manuscript review, after the original analyses.

\begin{table}[!htbp]
\centering
\begin{minipage}{0.8\textwidth}
\small
\caption{Retrospective omnibus tests that preserve task pairing. Adjusted probabilities use a family of four tests.}
\label{tab:paired-checks}
\begin{tabularx}{\linewidth}{@{}Xrr@{}}
\toprule
\tableheading{Comparison} & \tableheading{Permutation $p$} & \tableheading{Adjusted $p$} \\
\midrule
Model effect across four harnesses & 0.00005 & 0.00020 \\
Harness effect across four harnesses & 0.03775 & 0.15100 \\
Model effect across six harnesses & 0.00005 & 0.00020 \\
Harness effect across six harnesses & 0.03820 & 0.15280 \\
\bottomrule
\end{tabularx}
\end{minipage}
\end{table}

\begin{table}[!htbp]
\centering
\begin{minipage}{0.8\textwidth}
\small
\caption{Exact paired tests. Exclusive wins and losses count discordant tasks. Model and harness comparisons use separate correction families.}
\label{tab:exact-pairs}
\begin{tabularx}{\linewidth}{@{}Xrrr@{}}
\toprule
\tableheading{Comparison} & \tableheading{Wins} & \tableheading{Losses} & \tableheading{Adjusted $p$} \\
\midrule
sol baseline over luna baseline & 15 & 3 & 0.0226 \\
luna baseline over haiku baseline & 26 & 2 & $9.10\times10^{-6}$ \\
sol baseline over haiku baseline & 39 & 3 & $1.69\times10^{-8}$ \\
\midrule
sol react over baseline & 5 & 4 & 1.000 \\
sol plan execute over baseline & 3 & 8 & 1.000 \\
sol reflect retry over baseline & 11 & 6 & 1.000 \\
sol system heavy over baseline & 4 & 5 & 1.000 \\
sol notes memory over baseline & 3 & 12 & 0.527 \\
luna react over baseline & 3 & 7 & 1.000 \\
luna plan execute over baseline & 5 & 14 & 0.954 \\
luna reflect retry over baseline & 5 & 10 & 1.000 \\
luna system heavy over baseline & 4 & 8 & 1.000 \\
luna notes memory over baseline & 4 & 10 & 1.000 \\
haiku react over baseline & 2 & 11 & 0.337 \\
haiku plan execute over baseline & 6 & 10 & 1.000 \\
haiku reflect retry over baseline & 4 & 11 & 1.000 \\
haiku system heavy over baseline & 2 & 8 & 1.000 \\
haiku notes memory over baseline & 2 & 6 & 1.000 \\
\bottomrule
\end{tabularx}
\end{minipage}
\end{table}

\FloatBarrier

The wins and losses in Table~\ref{tab:exact-pairs} count tasks solved only by the first configuration and only by the second. Conditional on their sum, the exact two-sided binomial test assigns equal probability to either winner under the null. Model comparisons use a family of three baseline pairs. Harness comparisons use a separate family of 15 comparisons against the same model's baseline.

\end{document}